\documentclass[3p]{elsarticle}
\usepackage{hyperref}

\journal{Journal}

\usepackage{amsmath,amsthm}

\usepackage{graphicx}
\usepackage{algpseudocode}
\usepackage{algorithm}
\usepackage{algpascal}
\usepackage{bm}
\usepackage{hyperref}
\usepackage{amssymb}
\usepackage{xcolor}

\DeclareRobustCommand{\uvec}[1]{{%
		\ifcsname uvec#1\endcsname
		\csname uvec#1\endcsname
		\else
		\bm{\mathbf{#1}}%
		\fi
}}

\theoremstyle{plain}
\newtheorem{thm}{Theorem}[]

\newtheorem{prop}{Proposition}

\theoremstyle{definition}
\newtheorem{defn}{Definition}[]

\newtheorem{exmp}{Example}[section]

\theoremstyle{remark}
\newtheorem{rem}{Remark}[]

\newcommand{\makehighlightn}[1]{\textcolor{black}{#1}}

\begin{document}

\begin{frontmatter}

\title{Darboux-Frame-Based Parametrization for a Spin-Rolling Sphere on a Plane: A Nonlinear Transformation of Underactuated System to Fully-Actuated Model}


\author[mymainaddress]{Seyed Amir Tafrishi\corref{mycorrespondingauthor}}
\cortext[mycorrespondingauthor]{Corresponding author}
\ead{s.a.tafrishi@srd.mech.tohoku.ac.jp}

\author[Svinin]{Mikhail Svinin}

\author[mymainaddress2]{Motoji Yamamoto}

\address[mymainaddress]{Department of Robotics, Tohoku University, Sendai, Miyagi, Japan.}
\address[mymainaddress2]{Department of Mechanical Engineering, Kyushu University, Kyushu, Japan}
\address[Svinin]{Department of Information Science and Engineering, Ritsumeikan University, Kyoto, Japan}

\begin{abstract}
This paper presents a new kinematic model based on \makehighlightn{the Darboux frame} for motion control and planning. In this work, we show that an underactuated model of a spin-rolling sphere on a plane with five states and three inputs can be transformed into a fully-actuated one by \makehighlightn{a given Darboux frame} transformation. This nonlinear state transformation establishes a geometric model that is different from conventional state-space ones. First, a kinematic model of the Darboux frame at the contact point of the rolling sphere is established. Next, we propose a virtual surface that is trapped between the sphere and the contact plane. This virtual surface is used for generating arc-length-based inputs for controlling the contact trajectories on the sphere and the plane. Finally, we discuss the controllability of this new model. In the future, we will design a geometric path planning method for the proposed kinematic model.
\end{abstract}

\begin{keyword}
Darboux frame \sep parameterization \sep spin-rolling \sep sphere on plane \sep controllability \sep underactuated system
\end{keyword}

\end{frontmatter}


\section{Introduction}
Sphere is a unique geometrical object that can be visualized as a fingertip \cite{okamura2000overview,kiss2002motion,droukas2016rolling,Finger3dofkinematics,cui_sun_dai_2017}, a convex object \cite{sumer2008rolling,diller2013micro,fernandez2017three} or a rolling robot \cite{ishikawa2011volvot,RollRollerRobotThesis2014,Tafrishi2019,fankhauser2010modeling,spinjohnson2018fuzzy}. Path planning and control of this underactuated model is a well-known problem. However, a proper kinematic parametrization can fundamentally make the control problem  easier.

In the kinematics of the rolling contact, the pure-rolling motion has \makehighlightn{two} degrees of freedom whose instantaneous rotation axis is located at the contact point. This axis is always parallel to the common tangent plane of two surfaces e.g., sphere and plane. However, spin-rolling motion, also known as twist-rolling, has \makehighlightn{three} degrees of freedom with a similar instantaneous rotation axis that passes through contact point. Also, its axis can be in any arbitrary direction because the spinning motion is normal to the rolling axis.

From the physical point of view, spin-rolling motions can be realized in such mechanical systems as gimbals. Another example comes from grasping an object by robotic hands with fingers. If we model the fingertip of the hand as a hemisphere, the angular spinning can be achieved by wrist rotation \cite{Finger3dofkinematics}. Rotating the spherical object in multiple directions with control of the fingertips is sometimes called as dexterous manipulation  \cite{okamura2000overview,kiss2002motion,cui_sun_dai_2017}. Yet another example is mechanical actuators with \makehighlightn{three} or more degrees of freedom in realizing spin-rolling motions which are developed for spherical mobile robots \cite{ishikawa2011volvot,Tafrishi2019,tafrishi2019fluid,tafrishi2019dynamics} and Ballbots \cite{fankhauser2010modeling,spinjohnson2018fuzzy}. In particular, spherical robots can be actuated by multiple rotating masses or use different cart-based actuators inside their shells to create spin-rolling locomotion. In addition, rotating spheres have broad applications in the field of nano and micro manipulations \cite{sumer2008rolling,diller2013micro,fernandez2017three}.

The conventional pure rolling model, with the sphere having two degrees of freedom is often referred to as a ball-plate system \cite{jurdjevic1993geometry,marigo2000rolling}. Since the ball is sandwiched between the moving plane and the ground, the system binds with a spinning constraint. The ball-plate system does not fit the spin-rolling case when there is spinning around the perpendicular axis of the surface plane. Thus, our model excludes mechanisms that manipulate convex objects e.g., the sphere, by planes \cite{marigo2000rolling,NonprehensilePlanningPlane2019} since rotating the plane around its normal axis cannot spin the sphere physically. To deal with this issue, Kiss et al. \cite{kiss2002motion} proposed a kinematic model with three independent planes to manipulate the sphere and controlled the relative angles by ignoring the plane configuration. Also, Date et al. used the advantage of spinning in an indirect way \cite{date2004simultaneous}. Their control algorithm was iteratively shifting the coordinate of the actuating plane with respect to different base frames, which was, in a sense, utilizing a third virtual rotation center. However, time scaling with the included coordinate transformation of the kinematic model can result in uncontrollable states \cite{Oriolo2005Feedback}. Also, the practicality of this approach was not discussed for realizing the proposed virtual rotation center by different propulsion mechanisms.

On the other hand, the Montana kinematic model \cite{Montana1988} illustrated that spinning can be included in the rolling sphere on the surface. It is also clear that having one more input (spinning or twist) results in the five-by-three kinematic model, which increases the level of accessibility. However, the spinning of the rolling object creates a certain complexity because the spin angle changes all the rotational states. This new system leaves the conventional planning in geometric phase shifting \cite{jurdjevic1993geometry,planningli1990,mukherjee2002motion,svinin2008motion,bai2018dynamics} hard to be applied. In order to find a model that is easier to develop motion planning and control strategies, a different parametrization of the kinematics model can help to reduce the complexity of the problem. Recently, the Darboux-frame-based kinematics has seen serious attention because it is time- and coordinate-invariant \cite{cui_sun_dai_2017,CuiDarboux2010,CuiDarboux2015}. L. Cui and J. Dai presented the Darboux-frame-based kinematics on the spin-rolling sphere with point contact under the rolling constraint \cite{CuiDarboux2010}. They investigated the formulation based on \makehighlightn{the Darboux frame} in the form of polynomials for rolling loci of the instantaneous kinematics on different surfaces \cite{CuiDarboux2015}. However, these studies did not consider a generalized model for the control and planning problems.


In this paper, we design a novel kinematic model with two main motivations: developing a geometric kinematic model with inputs in the arc-length domain, and transforming the underactuated model into a fully-actuated and controllable one. First, we establish a geometric model where the control inputs are time- and coordinate-invariant. To derive the kinematic model with the arc-length-based inputs, the Darboux frame is defined at the contact point between the rotating object and the fixed surface. Then, a virtual surface is introduced in order to manipulate the inputs of the kinematic model. Our model is different from the one established  in \cite{cui_sun_dai_2017,CuiDarboux2010,CuiDarboux2015} since it uses a virtual surface that can control the transformed system in the arc-length domain. Also, the induced curvatures of surfaces (the sphere and the plane) are designed dependent on the spinning angles through the Darboux frame. These curvatures can be designed to constrain the virtual surface inputs. The resulting extra angular constraints let the spin-rolling sphere to be manipulated for reaching desired states on the fixed surface (plane). To the best of our knowledge, this is the first arc-length-based model that transfers the kinematic model to a more accessible one. Note that this geometric model separates the time scale from the kinematics, which allows driving the system with different convergence rates in a given time. Finally, we check the controllability of this new kinematic model. It is important to note that our designed formulation through the Darboux frame increases the number of input parameters to five, which makes the control problem easier.

This paper is organized as follows. In Section 2, the Darboux-frame-based kinematic model is derived and explained by considering curvature properties of surfaces. Then, we establish a fully-actuated kinematic model from our nonlinear transformation. Next, the controllability of the new kinematic model is studied in Section 3. Finally, we conclude our findings in Section 4.

\section{Kinematic Model of Moving Darboux Frame}
In this section, a new model of the Darboux frame at the contact point of the spin-rolling sphere and plane with an introduced virtual surface is developed at first. Next, the induced curvatures with relative angles between the sphere, plane, and the established Darboux frame are found. Finally, the Darboux frame kinematics are substituted into the Montana equations \cite{Montana1988} and a fully-actuated model for the sphere-plane system is obtained. As we derive the new geometric model, the kinematics readily suit both the contact trajectories and parameters of the surfaces \cite{CuiDarboux2010,DIfgeometry1976}. Also, this transformation provides three significant benefits. First, the spin-rolling angular rotations explicitly appear on the relative curvature and torsion that makes it easier for manipulation.  Second, the Darboux frame kinematic equations separate the time variable from the planning due to its time- and coordinate-invariance (arc-length-based control inputs). Third, the underactuated sphere-plane system can be transformed into a fully-actuated one to simplify the control and planning problems.
\begin{figure}[t!]
	\centering
	\includegraphics[width=3.4 in]{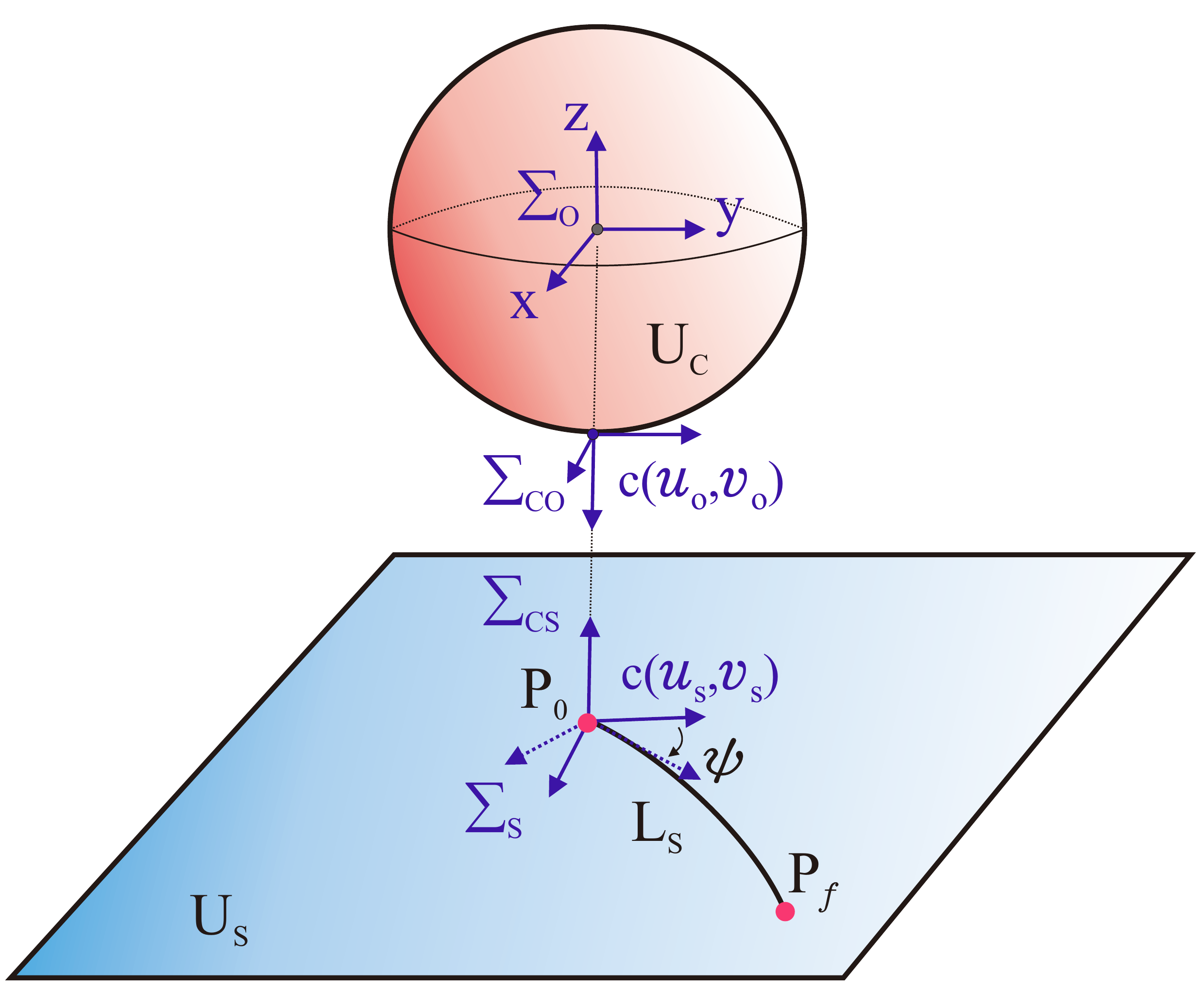}	
	\caption{Kinematic model of rotating sphere. Note: $\psi$ is the spin angle between sphere and plane surfaces.}\label{Fig:PlanningGeneralObject}
\end{figure}

\makehighlightn{Fig. \ref{Fig:PlanningGeneralObject} shows} a rotating object and fixed surface coordinate frames in the ball and plane system. $\Sigma_{o}$ and $\Sigma_{s}$ are the fixed frames on the rolling object (red sphere) and the plane. There are contact frames for a sphere $\Sigma_{co}$ and fixed traveled surface $\Sigma_{cs}$. Also, $\Sigma_{s}$ frame is fixed relative to other frames. It is assumed that the sphere with radius $R_o$ is rotating with no sliding constraint. The local coordinate systems for the sphere at $\Sigma_{co}$ and plane at $\Sigma_{cs}$ are defined \cite{Montana1988,planningli1990,svinin2008motion} as follows
\begin{align}
\begin{split}
&f_{o}: U_C \rightarrow \mathbb{R}^3
: c(u_{o},v_{o})\mapsto (-R_{o}\sin{u_{o}}\cos{v_{o}},R_{o}\sin{v_{o}},-R_{o}\cos{u_{o}}\cos{v_{o}}),\\
&f_{s}: U_S \rightarrow \mathbb{R}^3
: c(u_{s},v_{s}) \mapsto (u_{s},v_{s},0),
\end{split}
\label{Eq:Coorinatequationsphereonplane}
\end{align}
where $c(u_{o},v_{o}) \in [-\pi,\pi]$ and $c(u_{s},v_{s})$ are contact parameters of the sphere and plane. Coordinates $(u_{o},v_{o})$ are known to be the latitude and longitude of spherical surface $U_C$, respectively. Note that there is a spin angle $\psi$ between x-axis of $\Sigma_{co}$ sphere and $\Sigma_{cs}$ plane contact frames as Fig. \ref{Fig:PlanningGeneralObject}. For the sphere, denoted as object $U_C$, we have the following curvature characteristics \cite{planningli1990}
\begin{align}
\begin{split}
k^{o}_{nu}=k^{o}_{nv}=1/R_o,\;\tau^{o}_{gu}=\tau^{o}_{gv}=0,\; k^{o}_{gu}=\tan(v_{o})/R_{o},\;k^{o}_{gv}=0,
\label{Eq:detailMontanaCOonplane}
\end{split}
\end{align}
where $\{$$k^{o}_{nu}$, $\tau^{o}_{gu}$, $k^{o}_{gu}$$\}$ and $\{$$k^{o}_{nv}$, $\tau^{o}_{gv}$, $k^{o}_{gv}$$\}$ are the normal curvature, geodesic torsion and geodesic curvature of the rotating body with respect \makehighlightn{to $u_o$ and $v_o$}.
For the the plane surface $U_S$, the characteristics are
\begin{align}
\begin{split}
&k^{s}_{nu}=k^{s}_{nv}=\tau^{s}_{gu}=\tau^{s}_{gv}=k^{s}_{gu}=k^{s}_{gv}=0,
\label{Eq:MontanaCSNN}
\end{split}
\end{align}
where $\{$$k^{s}_{nu}$, $\tau^{s}_{gu}$, $k^{s}_{gu}$$\}$ and $\{$$k^{s}_{nv}$, $\tau^{s}_{gv}$, $k^{s}_{gv}$$\}$ are the normal curvature, geodesic torsion and the geodesic curvature of the fixed surface (plane) with respect \makehighlightn{to $u_s$ and $v_s$}.
\begin{figure}[t!]
	\centering
	\includegraphics[width=3.8 in]{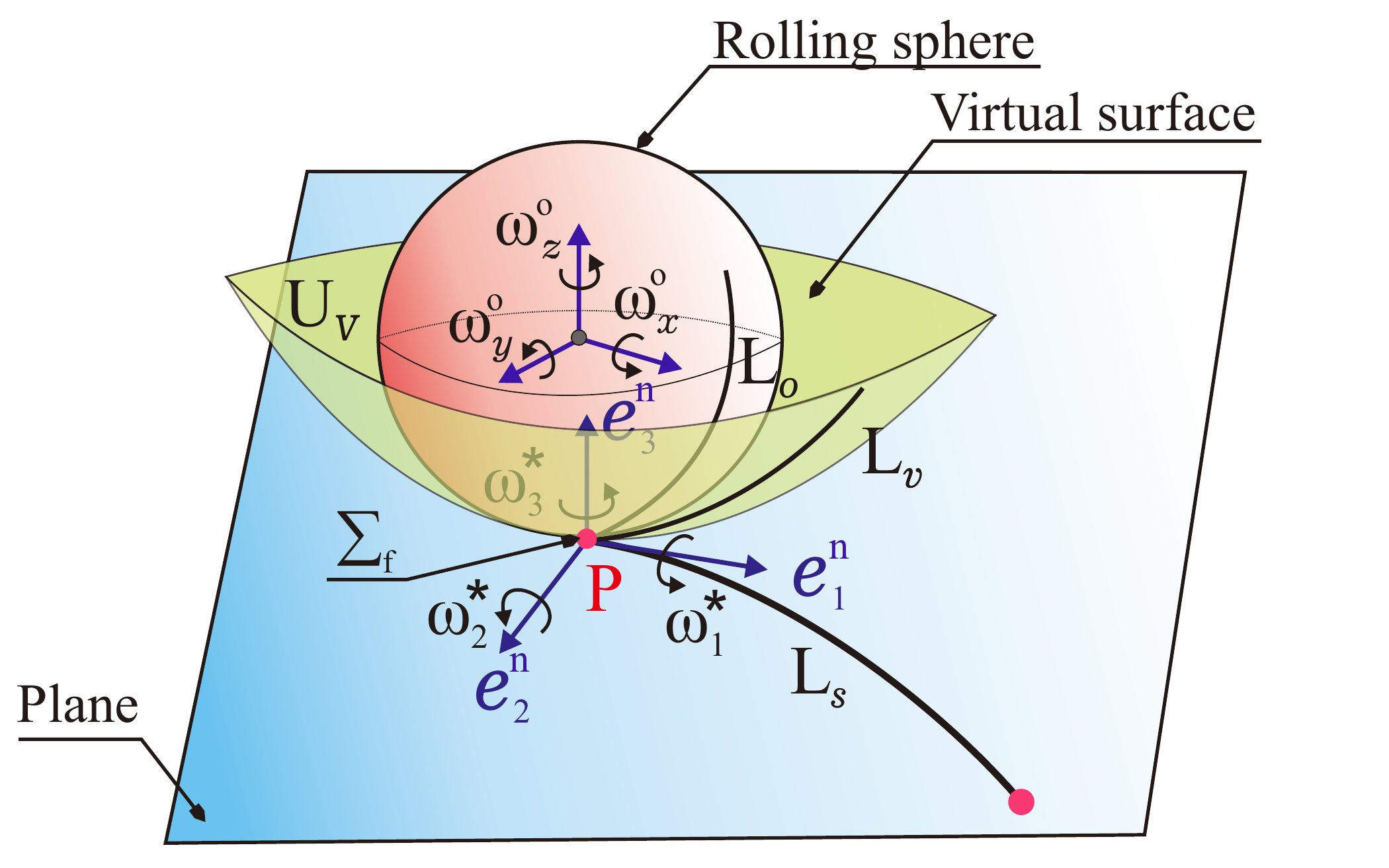}		
	\caption{Frame transformations of the spin-rolling sphere on a moving frame $\Sigma_f$. Note that $n$ superscript in $\{\bm{e}^n_1,\bm{e}^n_2,\bm{e}^n_3\}$ stands for the aligned Darboux frames of the sphere $(n=o)$, the plane $(n=s)$ and the virtual surface $(n=v)$. \label{Fig:Thecoordinatecontroltworotation}}
\end{figure}

To derive the kinematics in arc-length domain, we introduce a Darboux frame $\Sigma_{f}$ on the contact plane. Let a traced curve $\uvec{L}_{s}$ in Euclidean space be on the surface plane $U_S$ (see Fig. \ref{Fig:Thecoordinatecontroltworotation}). At the contact point $\uvec{P} \in U_S$, the unit vectors of the Darboux frame are ($\bm{e}^s_1,\bm{e}^s_2,\bm{e}^s_3$), where $\bm{e}^s_1$ is a tangent vector to the path $\uvec{L}_{s}$, $\bm{e}^s_3$ is a normal vector to the surface $U_S$ and $\bm{e}^s_2$ is perpendicular to the plane $\bm{e}^s_3 \times \bm{e}^s_1$ \cite{Riemannian2002} (see also \ref{MovingFramePreliminaries}).
To determine the angular velocity of the Darboux frame $\bm{\omega}^*$ attached to the plane trajectory $\uvec{L}_s$ at the contact point $\uvec{P}$, we
introduce two more frames, one, denoted as ($\bm{e}^o_1,\bm{e}^o_2,\bm{e}^o_3$) is attached to the sphere trajectory $\uvec{L}_o$ and
another one, denoted as ($\bm{e}^v_1,\bm{e}^v_2,\bm{e}^v_3$), is attached to the trajectory of a virtual
surface $\uvec{L}_v$ (see Fig. \ref{Fig:Thecoordinatecontroltworotation}). Note that all these frames coincide with each other \cite{CuiDarboux2010} due to no-sliding constraint.

\begin{defn}
The virtual surface $U_v$ (see Fig. \ref{Fig:Thecoordinatecontroltworotation} as an example) is characterized by the geodesic curvature $\alpha_s$, the geodesic torsion $\beta_s$, and the normal curvature $\gamma_s$. These variables $\{\alpha_s,\beta_s,\gamma_s\}$ will be replaced with the angular velocity inputs of the Montana Kinematic model \cite{Montana1988} upon transformation to the arc-length domain.
The virtual surface can be imagined as a sandwiched surface that has curvatures variations (geometric control inputs) projected onto both sphere and plane trajectories with resultant angular velocity $\bm{\omega}^*$ \footnote{Example \ref{Ex:examplerelationvirtual} shows this kinematic relation for a rolling disc case.}. As explained earlier, the Darboux frames of all three surfaces are always aligned with each other (see Fig. \ref{Fig:Thecoordinatecontroltworotation}). Since the sphere rolls on the plane, 
the arbitrary virtual surface at common contact point $\uvec{P}$ can be imagined as a flexible sandwiched sheet between the ball and plate whose relative arc length variation $ds$ results in the angular change of the spin-rolling sphere. Thus, the angular velocity of the Darboux frame along $\uvec{L}_s$
 at $\uvec{P}$ is established (see \ref{VirtualSurfaceDarbouxSurfaces}) as
\begin{align}
\bm{\omega}^*= \delta  (-\tau^{*}_g \bm{e}^s_1+k^{*}_n\bm{e}^s_2-k^{*}_g\bm{e}^s_3),
\label{Eq:angularveloctytotal}
\end{align}
where
\begin{eqnarray}
\delta=ds/dt,\;k^{*}_g= k^{o}_g-k^{s}_g-\alpha_s,\;k^{*}_n=k^{o}_n-k^{s}_n-\gamma_{s},\; \tau^{*}_g=\tau_g^{o}-\tau^{s}_g-\beta_{s},
\label{EQ:TheCurvaturerelevantDif}
\end{eqnarray}
where $\delta$ is the derivative of the arc length relative to time (the rolling rate), and $k^{*}_g$, $k^{*}_n$ and $\tau^{*}_g$ are, respectively, the induced geodesic curvature, the normal curvature and the geodesic torsion between two surfaces (sphere and plane) including the virtual surface's characteristics $\{$$\alpha_s$, $\beta_{s}$, $\gamma_{s}$$\}$. Also, $\{k^o_n,k^o_g,\tau^o_g\}$ and $\{k^s_n,k^s_g,\tau^s_g\}$ are the induced curvature (differential characteristics) of the sphere and plane with respect to the Darboux frame $\Sigma_f$.
\end{defn}

\begin{rem}
Since we want to parameterize the kinematic model by using Darboux frame and have its advantages \cite{cui_sun_dai_2017,CuiDarboux2010,CuiDarboux2015,DIfgeometry1976}, the defined virtual surface should exist in the formulation (\ref{Eq:angularveloctytotal}) for the ability to control the sphere-plate system in the arc-length domain (angle/arc length) by $\{$$\alpha_s$, $\beta_{s}$, $\gamma_{s}$$\}$ rather than the angular velocity in the time domain, presented in Montana kinematics \cite{Montana1988}. Without our defined virtual surface, the sphere-plane kinematics (\ref{Eq:angularveloctytotal}) will only correspond to certain constrained trajectories from the induced curvature of the geometric surfaces with invariant variables which were already presented in \cite{CuiDarboux2010}.
\end{rem}
\begin{rem}
In \cite{CuiDarboux2015}, the angular velocity of the Darboux frame (\ref{Eq:angularveloctytotal}-\ref{EQ:TheCurvaturerelevantDif}) had only similar $\delta \alpha_s$ term that it was defined as the compensatory spin rate. However, we have derived a general formulation with the defined arbitrary virtual surface to control the pure-rolling velocities of the sphere by $\{\delta\gamma_{s},\delta\beta_{s}\}$ and its spin velocity by $\delta \alpha_s$. Note that our virtual surface does not fit the proposed analytical solution in \cite{CuiDarboux2015} since the problem becomes unsolvable with unknown variables more than the polynomial formulations \makehighlightn{(five unknown variable and three equations)}.

\end{rem}

\begin{exmp}
	\label{Ex:examplerelationvirtual}
	In order to show the relation of the virtual surface inputs with angular velocity in a simple example, we consider a rolling disc case of radius $R$. Let the disc rolls with angle of $v_o$ without spinning angle which has the curvature of $k^o_g=\tau^o_g=0$ and $k^o_n=1/R$. The fixed surface $U_S$ is considered as a plane with $k^s_g=\tau^s_g=k^s_n=0$. Then, by substituting these curvatures into (\ref{Eq:angularveloctytotal}), we have the angular $\bm{\omega}^*$ and linear $\uvec{v}^*$ velocities at $\uvec{P}$ as follows
	\begin{equation}
	\bm{\omega}^*= \delta \left[(1/R)+\gamma_s\right]\bm{e}^s_2,\; \uvec{v}^*= \bm{\omega}^* \times \uvec{r}_s= \delta \left[(1/R)+\gamma_s\right]\bm{e}^s_2 \times R\;\uvec{e}^s_3 =\delta \left[1+ R \; \gamma_s  \right]\uvec{e}^s_1.
	\label{Eq:Rollingdiscunitbased}
	\end{equation}
	We can see that the curvature variation of $\gamma_s$ is projected onto the unit frames, which changes the linear and angular velocities. The disc travels on the plane along $\uvec{e}^s_1$ with corresponding kinematics (\ref{Eq:Rollingdiscunitbased}). Also, the kinematics (\ref{Eq:Rollingdiscunitbased}) can be represented in the conventional Cartesian coordinate $\{\uvec{s}-\uvec{i}\uvec{j}\uvec{k}\}$ system 
	as
	\begin{equation}
	 \uvec{e}^s_1=\uvec{i},\;\;\uvec{e}^s_2= \uvec{j},\;\; \delta =\frac{ds}{dt}= R  \frac{d v_o }{dt}.
	 \label{Eq:examplecasecartesianveldar}
	\end{equation}
	Note that (\ref{Eq:Rollingdiscunitbased})-(\ref{Eq:examplecasecartesianveldar}) can be extended for the spin-rolling system with three arc-length-based inputs.

\end{exmp}
\begin{figure}[t!]
	\centering
	\includegraphics[width=3 in]{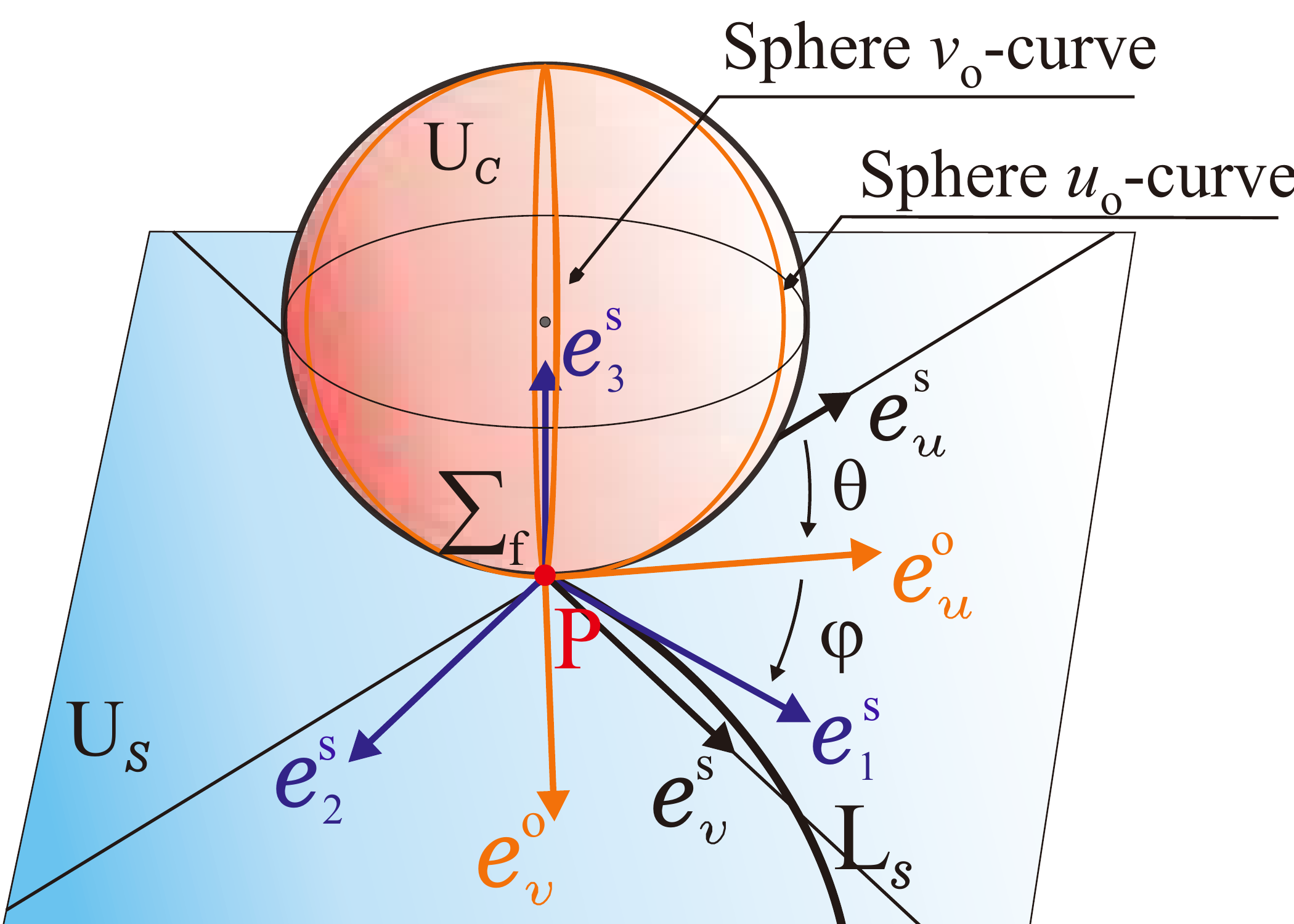}		
	\caption{The relation of the plane's Darboux frame ($\bm{e}^s_1,\bm{e}^s_2,\bm{e}^s_3$) along the trajectory $\uvec{L}_s$ with the induced contact point frames of the sphere ($\bm{e}^o_u,\bm{e}^o_v,\bm{e}^o_3$) and the plane ($\bm{e}^s_u,\bm{e}^s_v,\bm{e}^s_3$). }\label{Fig:ThecoordinateInducedCurvature}
\end{figure}

Now, we want to design the induced curvatures of the sphere $\{k^o_n,k^o_g,\tau^o_g\}$ and plane $\{k^s_n,k^s_g,\tau^s_g\}$ for (\ref{Eq:angularveloctytotal}-\ref{EQ:TheCurvaturerelevantDif}) in a way that they become constrained by angles between two surfaces and the Darboux frame. There are the orthonormal frames of ($ \bm{e}^o_u$, $\bm{e}^o_v$, $\bm{e}^o_3 $) and ($\bm{e}^s_u$, $\bm{e}^s_v$, $\bm{e}^s_3$) (
see Fig.~\ref{Fig:ThecoordinateInducedCurvature}) induced by the contact coordinates of the rotating object (sphere) $\Sigma_{co}$ and the fixed surface (plane) $\Sigma_{cs}$, respectively. \makehighlightn{These vectors present the contact} point's tangential velocity  of each surface that can be found by taking the derivative of local coordinates in (\ref{Eq:Coorinatequationsphereonplane}). Note that both frames of the contact point's velocities of the sphere ($ \bm{e}^o_u$, $\bm{e}^o_v$) and the plane ($\bm{e}^s_u$, $\bm{e}^s_v$) reside on the plane of the Darboux frame ($\bm{e}^s_1,\;\bm{e}^s_2$) of the curve $\uvec{L}_s$. Additionally, the unit normals of the surfaces ($\bm{e}^o_3 $ and $\bm{e}^s_3$) and the Darboux frame of the plane are always aligned with each other.

\makehighlightn{%
Let $\varphi$ be the rotation angle around vector $\bm{e}_3^{s}$. This angle is between tangent vector to $u_o$-curve, $\bm{e}^{o}_u$, and the
Darboux frame's vector $\bm{e}^s_1$ that is tangent to path $\uvec{L}_s$ as shown in Fig.~\ref{Fig:ThecoordinateInducedCurvature}.
}
Also, let $\theta$ be the angle (of rotation about $\bm{e}_3$) between the vectors of the sphere $\bm{e}^{s}_u$ and plane $\bm{e}^{o}_u$ contact frames. The motion equations of the Darboux frame with respect to the induced coordinates are summarized in \ref{MovingFramePreliminaries}, which was also shown in \cite{cui_sun_dai_2017,CuiDarboux2010,Riemannian2002}.
Then, the induced curvature between the Darboux frame $\Sigma_f$ and each of the contact coordinates of the sphere  $\Sigma_{co}$ and the plane $\Sigma_{cs}$ can be developed (see  \ref{DarbouxFrameKinematicsSec}). By using the established relation (\ref{Eq:MainCurvatureDarobux}), the normal curvature $k^{o}_n$, the geodesic curvature $k^{o}_g$ and geodesic torsion $\tau^{o}_g$ of the sphere in the direction of $\bm{e}^s_1$ can be found in terms of the angle $\varphi$ as follow 
\begin{eqnarray}
k^{o}_n&=&k^{o}_{nu}\cos^2\varphi+2\tau^{o}_{gu}\cos\varphi\sin\varphi+k^{o}_{nv}\sin^2\varphi = 1/R_o,\nonumber\\
\tau^{o}_g&=&\tau^{o}_{gu}\cos 2\varphi+\frac{1}{2}(k^{o}_{nv}-k^{o}_{nu})\sin 2\varphi=0,\nonumber\\
k^{o}_g&=&k^{o}_{gu}\cos\varphi+k^{o}_{gv}\sin\varphi=\tan v_{o}\cos\varphi/R_{o}.
\label{Eq:MainCurvatureDarobuxCO}
\end{eqnarray}
The normal curvature $k^{s}_n$, the geodesic curvature $k^{s}_g$ and the geodesic torsion $\tau^{s}_g$ of the plane in the direction of $\bm{e}^s_1$ vecotr ($\bm{e}^s_u-\bm{e}^s_1$ make the angle of $\varphi+\theta$ as shown in Fig.~\ref{Fig:Thecoordinatecontroltworotation}) is
\begin{eqnarray}
k^{s}_n&=&k^{s}_{nu}\cos^2(\theta+\varphi)+2\tau^{s}_{gu}\cos(\theta+\varphi)\sin(\theta+\varphi)+k^{s}_{nv}\sin^2(\theta+\varphi)=0, \nonumber\\
\tau^{s}_g&=&\tau^{s}_{gu}\cos 2(\theta+\varphi)+\frac{1}{2}(k^{s}_{nv}-k^{s}_{nu})\sin 2(\theta+\varphi)=0,\nonumber\\
k^{s}_g&=&k^{s}_{gu}\cos(\theta+\varphi)+k^{s}_{gv}\sin(\theta+\varphi)=0.
\label{Eq:MainCurvatureDarobuxCS}
\end{eqnarray}
Note that the derived induced curvatures of the sphere (\ref{Eq:MainCurvatureDarobuxCO}) and plane (\ref{Eq:MainCurvatureDarobuxCS}) will be substituted into (\ref{EQ:TheCurvaturerelevantDif}) for finding the angular velocity of the Darboux frame $\bm{\omega}^*$.

Eq. (\ref{Eq:angularveloctytotal}) is the angular velocity of the Darboux frame. Transformation that expresses the Darboux frame ($\bm{e}^s_1,\bm{e}^s_2,\bm{e}^s_3$) in the frame ($\bm{e}^{o}_u,\bm{e}^{o}_v,\bm{e}^{o}_3$) of the sphere is
\begin{align}
\begin{split}
&\bm{e}^s_1=\cos(\varphi+\theta)\bm{e}^o_u+\sin(\varphi+\theta)\bm{e}^o_v,\\
&\bm{e}^s_2=-\sin(\varphi+\theta)\bm{e}^o_u+\cos(\varphi+\theta)\bm{e}^o_v,\\
&\bm{e}^s_3=\bm{e}^o_3.
\end{split}
\label{Eq:TransformationDarbSph}
\end{align}
The angular velocity of the Darboux frame $\bm{\omega}^*$ is equal to the angular velocity of the sphere. Thus, one gets the angular velocity of the sphere $\bm{\omega}^o$ by substituting (\ref{Eq:TransformationDarbSph}) into (\ref{Eq:angularveloctytotal}) (see Fig.~\ref{Fig:Thecoordinatecontroltworotation}) as follows
\begin{eqnarray}
\bm{\omega}^{o}=\omega^{o}_x \bm{e}^o_u +\omega^{o}_y \bm{e}^o_v
+\omega^{o}_z \bm{e}^o_3,
\label{Eq:angularveloctytotalMAIN}
\end{eqnarray}
where
\begin{align}
\begin{split}
&\omega^{o}_x=\delta(-\cos{(\varphi+\theta)}\tau^{*}_g-\sin{(\varphi+\theta)}k^{*}_n),\\
& \omega^{o}_y=\delta(-\sin{(\varphi+\theta)}\tau^{*}_g+\sin{(\varphi+\theta)}k^{*}_n),\\
&\omega^{o}_z=\delta(-k^{*}_g).
\end{split}
\label{EQ:AngularDATA}
\end{align}
The angular velocities (\ref{EQ:AngularDATA}) are formulated in terms of $\delta$, $\theta$ and $\varphi$, including the virtual surface where $\delta$ is the rolling rate of the sphere and $\theta$ and $\varphi$ angles are for assigning the direction of the sphere on the plane. Because we use the kinematics that gives the angular velocities ($ \omega^{o}_x,\omega^{o}_y,\omega^{o}_z$), all these input variables directly change the motion of the sphere on the plane.

The angular velocities of the Darboux frame in (\ref{EQ:AngularDATA}) have to be transferred to the states of the sphere and plane for obtaining the transformation of the fully-actuated model. We utilize the Montana equation \cite{Montana1988} with the inclusion of no-sliding constraints. By knowing the curvatures of the surfaces (\ref{Eq:detailMontanaCOonplane}-\ref{Eq:MontanaCSNN}), we write down the Montana equations as follows \cite{Tafrishi2019,marigo2000rolling}
\begin{equation}
	\begin{split}
	\left[\begin{array}{c}
	\dot{u}_{s}(t)\\
	\dot{v}_{s}(t)\\
	\dot{u}_{o}(t)\\
	\dot{v}_{o}(t)\\
	\dot{\psi}(t)
	\end{array}\right]=\left[\begin{array}{ccc}
	0& R_{o}&0\\
	- R_{o}&0&0\\
	-\sin{\psi}/\cos{{v}_{o}}&-\cos{\psi}/\cos{{v}_{o}}&0\\
	-\cos{\psi}&\sin{\psi}&0\\
	-\sin{\psi}\tan{{v}_{o}}&-\cos{\psi}\tan{{v}_{o}} &-1
	\end{array}\right]\left[\begin{array}{c}
	\omega^{o}_x\\
	\omega^{o}_y\\
	\omega^{o}_z
	\end{array}\right], \\
	\end{split}
\label{Eq:LastMonata2D}
\end{equation}
where $\psi$ is the spin angle between the sphere and plane (see Fig.~\ref{Fig:PlanningGeneralObject}).
The angular velocity of the spin-rolling sphere $\bm{\omega}^o$ in our Darboux frame definition in (\ref{EQ:AngularDATA}) is constructed by including the sphere and plane induced curvatures in (\ref{Eq:MainCurvatureDarobuxCO}-\ref{Eq:MainCurvatureDarobuxCS}) and the virtual surface. Finally, we substitute
(\ref{EQ:AngularDATA}) into ($\omega^{o}_{x},\omega^{o}_{y},\omega^{o}_{z}$) inputs of (\ref{Eq:LastMonata2D}), which results in
\begin{align}
\begin{split}
&\left[\begin{array}{c}
u'_{s}(s)\\
v'_{s}(s)\\
u'_{o}(s)\\
v'_{o}(s)\\
\psi'(s)
\end{array}\right]=\left[\begin{array}{c}
\sin(\theta+\varphi)\\
\sin(\theta+\varphi)\\
\frac{\sin(\theta+\varphi)[\sin{\psi}-\cos{\psi}]}{R_{o}\cos{v_{o}}}\\
\frac{\sin(\theta+\varphi)[\cos{\psi}+\sin{\psi}]}{R_{o}}\\
\frac{\tan{{v}_{o}}[\sin(\theta+\varphi)(\sin\psi-\cos\psi)+\cos\varphi]}{R_{o}}
\end{array}\right]+\left[\begin{array}{c}
0\\
0\\
0\\
0\\
-1
\end{array}\right]\;\alpha_s\\
&+
\left[\begin{array}{c}
R_{o}\sin(\theta+\varphi)\\
-R_{o}\cos(\theta+\varphi)\\
\frac{-\sin{(\psi+\theta+\varphi)}}{\cos{v_{o}}}\\
-\cos{(\psi+\theta+\varphi)}\\
-\tan{{v}_{o}}\sin{(\psi+\theta+\varphi)}
\end{array}\right]\;\beta_{s}
+
\left[\begin{array}{c}
-R_{o}\sin(\theta+\varphi)\\
-R_{o}\sin(\theta+\varphi)\\
\frac{\sin(\theta+\varphi)[\cos{\psi}-\sin{\psi}]}{\cos{v_{o}}}\\
-\sin(\theta+\varphi)[\sin{\psi}+\cos{\psi}]\\
\tan{{v}_{o}}[\sin(\theta+\varphi)(\cos\psi-\sin\psi)]
\end{array}\right]\;\gamma_{s}.
\end{split}
\label{EQ:LatestStateEquation}
\end{align}
We now describe the new kinematic model in following propositions.
\begin{prop}
	\label{DriftTermAnglethetaandVarphi}
	The Darboux-frame-based kinematics (\ref{EQ:LatestStateEquation}) is the components of $\{\theta$, $\varphi \}$. The main function of the introduced $\{\theta,\varphi\}$ angles are to constrain virtual surface inputs $\{\alpha_s,\beta_s,\gamma_s\}$. It is assumed that the sphere path on the plane $\uvec{L}_s$ is known and the angular constraint $\{\theta,\varphi\}$ keeps the spin-rolling sphere always in the assigned angular direction $G_f$ on the plane $U_S$ by
	\begin{align}
	&\theta(\beta_s,\gamma_s,G_f)=\cot^{-1} \Big [\frac{1}{\beta_{s}}\Big(
	\frac{1}{R_{o}}(1-\tan G_f) +\gamma_{s}(-1+\tan
	G_f)-\beta_{s}\tan{G_f} \Big) \Big], \nonumber\\
	&  \varphi(G_f)=\begin{cases}
	&\pi, \;\;\;\;\;\;\;\;\;\;\;-\frac{3\pi}{4}<G_f<0 \;\;\;\;\&\;\;\;\;\;\; 0 \leq G_f<\frac{\pi}{4}\\
	&0,\;\;\;\;\;\;\;\;\;\;\;\;-\pi<G_f<-\frac{3\pi}{4} \;\& \;-\pi \leq G_f<\frac{\pi}{4}
	\end{cases}.
	\label{Eq:AngleGoalthetaM}
	\end{align}
	By using Darboux-frame-based kinematics (\ref{EQ:LatestStateEquation}) with (\ref{Eq:AngleGoalthetaM}), the system has an independent angular input $G_f$ to converge desired plane states $\{u_{s,f},v_{s,f}\}$ and three arc-length-based inputs $\{\alpha_s,\beta_s,\gamma_s\}$ for controlling the spin-rolling states toward the desired angular states $\{$$u_{o,f}$, $v_{o,f}$, $\psi_f\}$.
\end{prop}
\begin{proof}
 To move in the direction of desired arbitrary angle $G_f$ with the assigned trajectory of $\uvec{L}_s$, we derive $\theta$ from  (\ref{EQ:LatestStateEquation}). The derivation is done by the definition of the goal angle $G_f$ on plane in a small size arc length step $d s$ by using the differential equations for $(u'_{s},v'_{s})$ in (\ref{EQ:LatestStateEquation})
\begin{equation}
\begin{split}
&G_f \overset{\Delta}{=}\tan^{-1}\left(\frac{v'_{s}}{u'_{s}}\right)= \tan^{-1}\left(\frac{d v_{s}}{d u_{s}}\right)=\tan^{-1}\Bigg ( \frac{\sin(\theta+\varphi)-R_{o}\sin(\theta+\varphi)\gamma_{s}-R_{o}\cos(\theta+\varphi)\beta_{s}}{\sin(\theta+\varphi)-R_{o}\sin(\theta+\varphi)\gamma_{s}+R_{o}\sin(\theta+\varphi)\beta_{s}}\Bigg ).
\end{split}
\label{Eq:Gfdefinitionmain}
\end{equation}
After factoring the numerator and denominator by $\sin(\theta+\varphi)$ and finding equation (\ref{Eq:Gfdefinitionmain}) for $\theta+\varphi$, we have
\begin{align}
\begin{split}
&\theta+\varphi=\cot^{-1} \Big [\frac{1}{\beta_{s}}\Big (
\frac{1}{R_{o}}(1-\tan G_f)+\gamma_{s}(-1+\tan
G_f)-\beta_{s}  \tan {G_f} \Big ) \Big ].
\end{split}
\label{Thetaupdatelastproof}
\end{align}
 \begin{figure}[t!]
	\centering	
	\includegraphics[width=4.4 in]{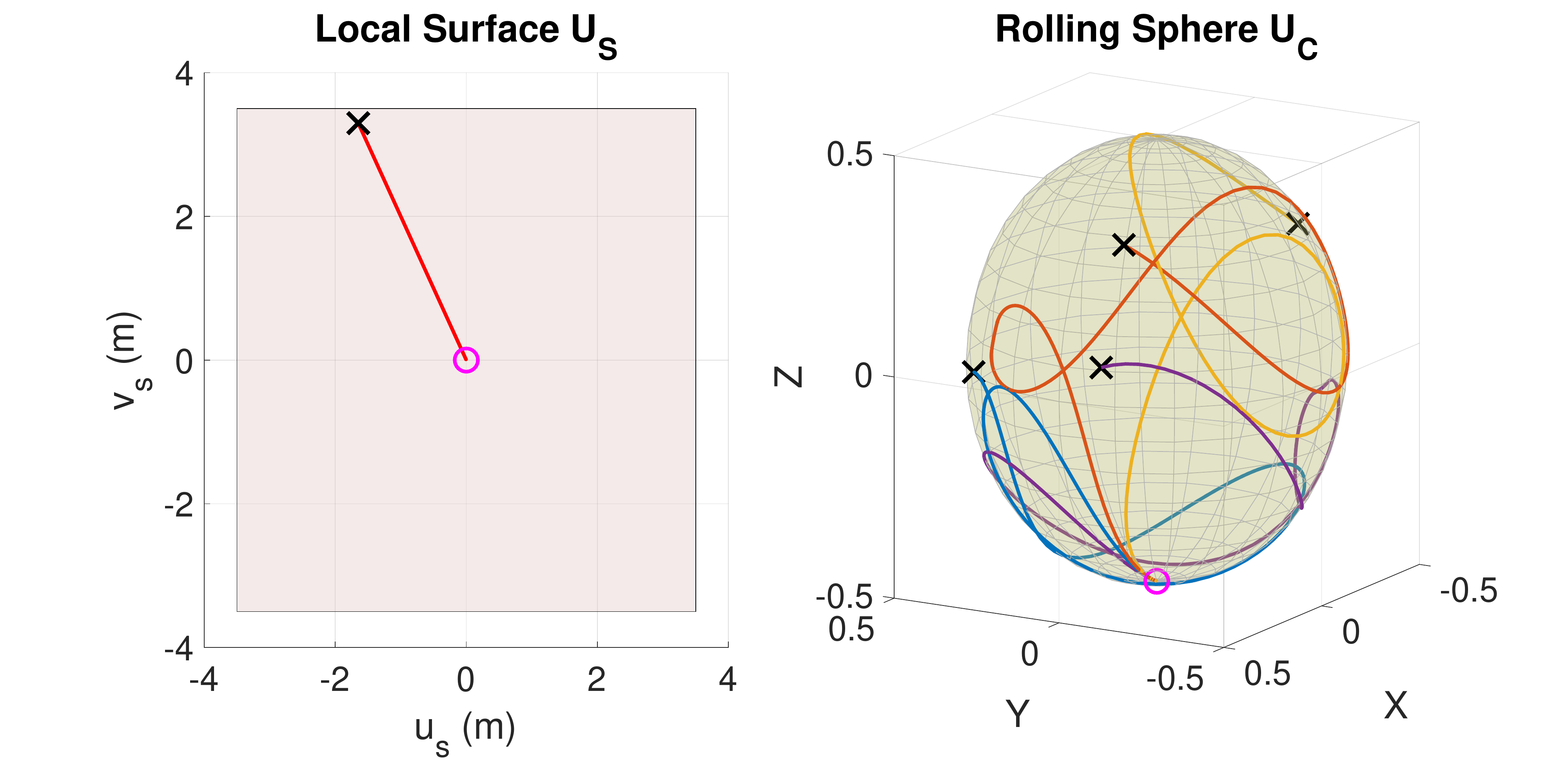}		
	\caption{Example results for the same constant goal $G_f$ on the plane but different arc-length inputs $\{\alpha_s,\beta_s,\gamma_s\}$.}\label{Fig:DarbouxAngleTheta1}
\end{figure}
Since $\theta$ is the angle between two surfaces' contact curves $\bm{e}^{o}_u-\bm{e}^{s}_u$, (\ref{Thetaupdatelastproof}) is assigned to angle $\theta$. Next, angle $\varphi$ only corresponds to angular input $G_f$  on the plane. Because we have the cotangent function in (\ref{Thetaupdatelastproof}), there is $\pi$ shift (the shifting line is located on $u_{s}=v_{s}$ of the plane as $G_f=\pm\pi/4$) in $\varphi$ depending on the desired angular input $G_f$ value to converge in all directions on the plane as (\ref{Eq:AngleGoalthetaM}). Based on the defined functions for $\theta$ and $\varphi$ angles, equation (\ref{Eq:AngleGoalthetaM}) is developed.
\end{proof}
\begin{rem}
	The nonlinear transformation has changed the underactuated $5\times3$ model (\ref{Eq:LastMonata2D}) to  $5\times4$ kinematic model in $s$-domain  (\ref{EQ:LatestStateEquation}) because $\{\theta(\beta_s,\gamma_s,G_f),\varphi(G_f)\}$ with angular input $G_f$ will always converge the system toward the desired plane states $\{u_{s,f},v_{s,f}\}$ and we have three arc-length-based inputs $\{\alpha_s,\beta_s,\gamma_s\}$ to converge the remaining desired angular states of the sphere $\{u_{o,f},v_{o,f},\psi_f\}$. \makehighlightn{However, we have one more input in the time-domain as the rolling rate $\delta$ (it comes from our definition in (\ref{Eq:angularveloctytotal}) providing the rest-to-rest motion of the sphere (similar to a time-scaling variable in  \cite{date2004simultaneous,sampei1986time})}. Thus, the transformed kinematic model in time domain becomes $5\times5$ system with $\{\alpha_s,\beta_s,\gamma_s,G_f, \delta\}$ inputs. If we look carefully, we can see that this nonlinear transformation lets us have $\{G_f, \delta\}$ new extra inputs. These two inputs are similar to a steering angular input and vehicle's input velocity in the kinematics of the 2-DoF wheeled mobile systems. 
We have also proved that one cannot obtain extra angular control input $G_f$ by directly using the Montana kinematics in \ref{Sec:Montanalimitationproblem}.
\end{rem}

\begin{rem}
	Fig. \ref{Fig:DarbouxAngleTheta1} provides a simulation example for a constant $G_f=\tan^{-1}\left[(v_{s,f}-v_{s,0})/(u_{s,f}-u_{s,0})\right]$ and a desired straight path on plane. The arc-length-inputs $\{\alpha_s,\beta_s,\gamma_s\}$ are changed while $\{\theta,\varphi\}$ angles constrain these inputs. This results in different trajectories on the sphere $U_C$ but the path on the plane $\uvec{L}_s$ is kept the same.
\end{rem}

\begin{figure}[t!]
	\centering	
	\includegraphics[width=4.4 in]{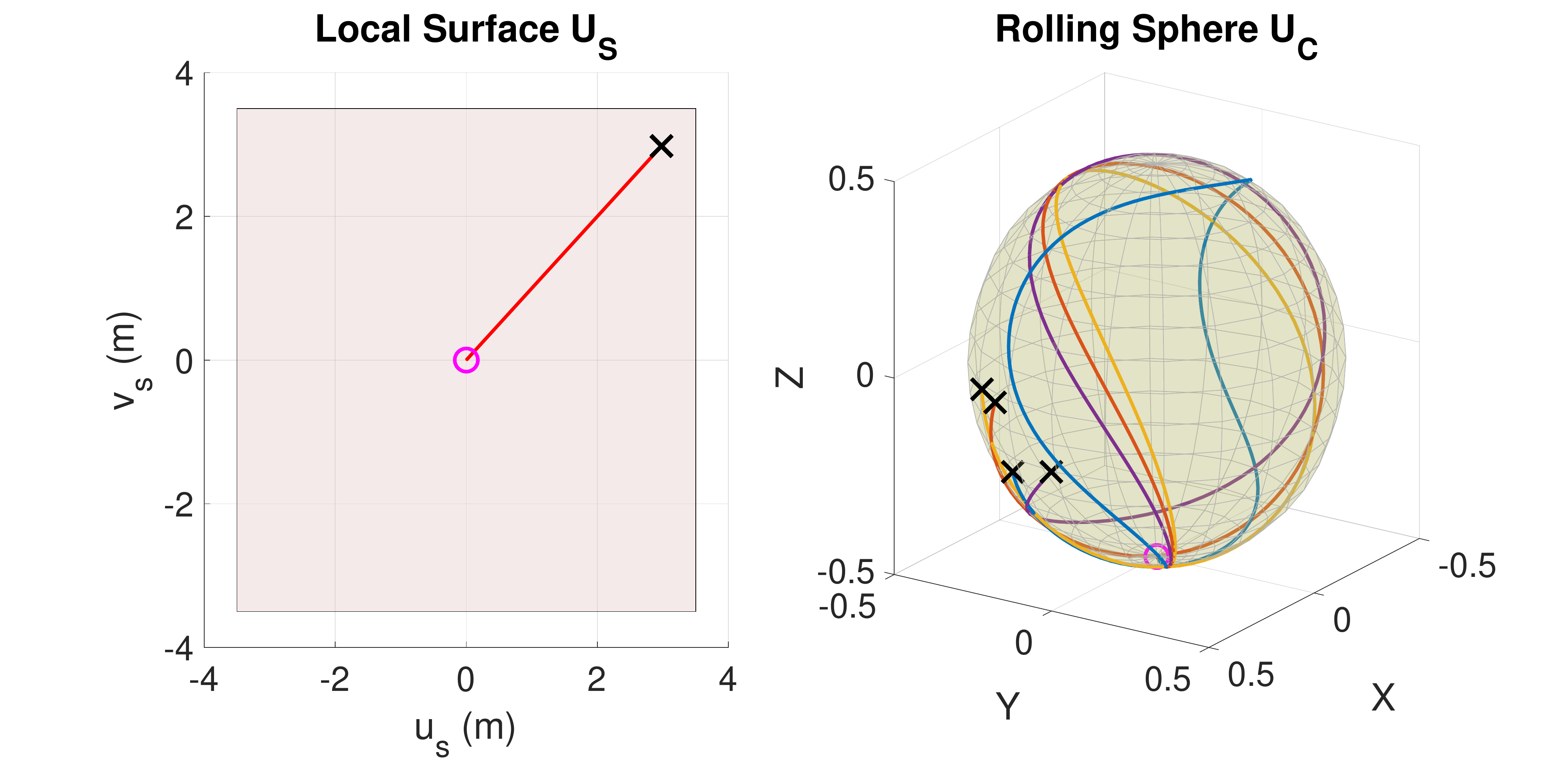}
	\includegraphics[width=4.4 in, height = 2.4 in]{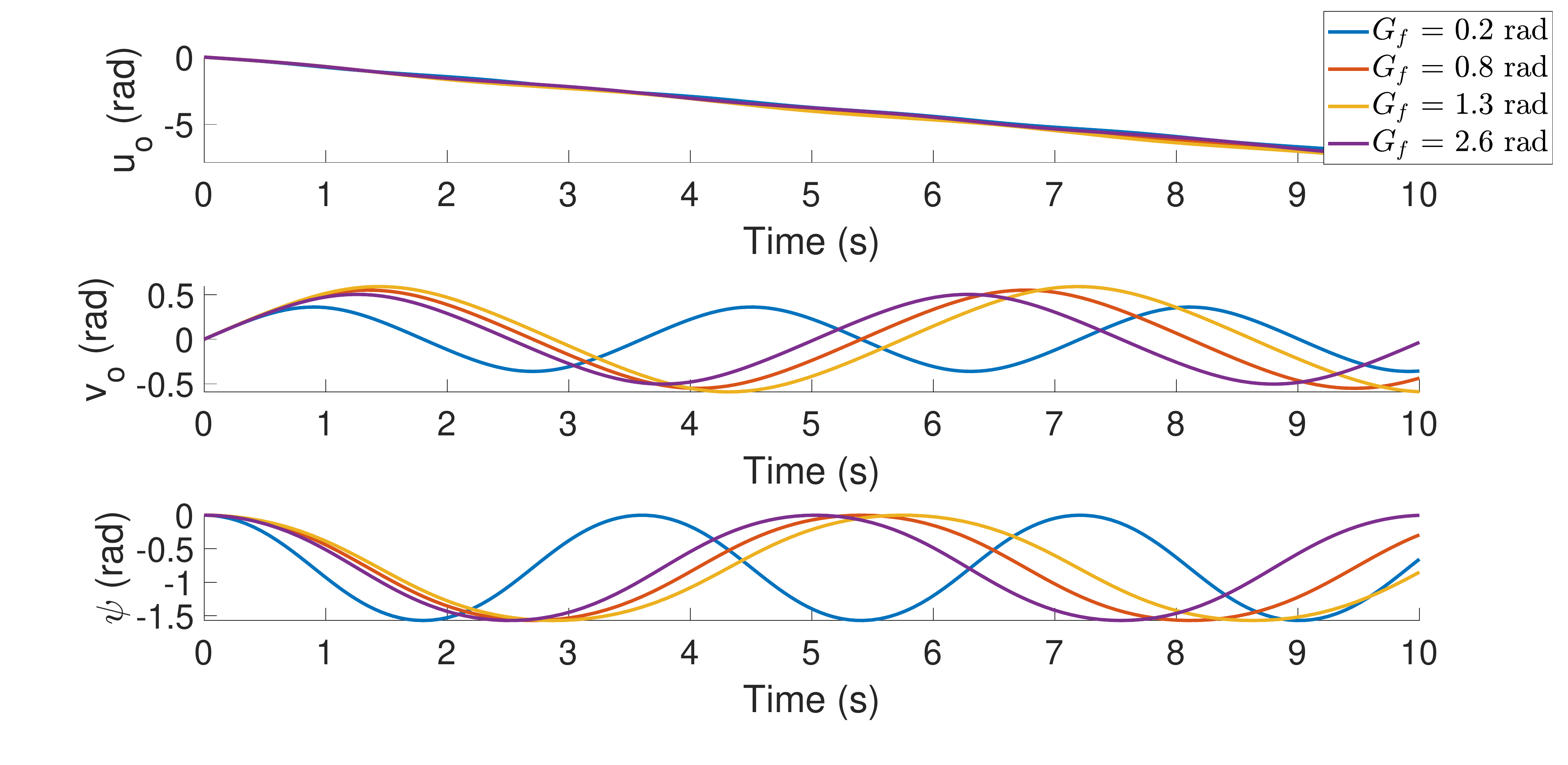}		
	\caption{The simulation analysis of the drift term for determining its effect on the ball-plate system.}\label{Fig:Dahetadrift1}
\end{figure}

The new kinematic model features a drift term. Geometrically, if we set the virtual surface inputs ($\alpha_s$, $\beta_{s}$, $\gamma_{s}$) zero, the drift term varies (change in $\{\theta,\varphi\}$ angles) the sphere velocity with constant angular direction on the plane ($45^o$) as shown in
Fig.~\ref{Fig:Dahetadrift1}. This drift term appears in the angular displacement as a sine wave with a constant periodic cycle. Note that the drift term appears due to the sphere's normal and geodesic curvatures (\ref{Eq:detailMontanaCOonplane}) which are dependent on $\{\theta,\varphi\}$
in (\ref{EQ:AngularDATA}). Note also that this drift term vanishes in the time-domain because the rolling rate $\delta$ that provides rest-to-rest locomotion converges to zero.

\section{Controllability of the Darboux-Frame-Based Kinematic Model}
\label{ControllabilitDarboux-BaseKinematic}
We now check the controllability of the kinematic model (\ref{EQ:LatestStateEquation}). This model can be represented as
\begin{align}
\dot{\uvec{x}}=\uvec{f}(x)+\sum_{i=1}^{3}\uvec{g}_i(x) \uvec{u}_i,
\label{generalsystems}
\end{align}
where $\uvec{f}(x)$ and $\uvec{g}_i(x)$ are our drift term and the control input coefficients; $\uvec{x}$=$\{u_{s}$, $v_{s}$, $u_{o}$, $v_{o}$, $\psi\}$ and  $\uvec{u}_i=\{\alpha_s,\beta_{s},\gamma_{s}\}$.
\begin{thm}
	\label{Controllabilitydrifttheorem}
	System (\ref{generalsystems}) is controllable if $\uvec{f}(x)$ is weakly positively Poisson stable (WPPS) and Lie algebra rank condition (LARC) is satisfied for local accessibility \cite{lian1994controllability,wang2011note}.
	\label{theoremofWPPS}
\end{thm}
In order to prove the weakly positively Poisson stable (WPPS) condition, we find whether the volume of phase space inside the given vector field for the drift term  $\uvec{f}(x)$ is preserved by Liouville's theorem \cite{arnol2013mathematical}:
\begin{align}
\nabla \cdot  \uvec{f}(x) = \sum_{i=1}^{5} \frac{\partial f^i}{\partial x^i}=\delta \sin(\theta+\varphi)(\sin\psi+\cos\psi)\frac{\tan{{v}_{o}}}{R_{o}}=0.
\label{EQ:LiouvilleTheorem}
\end{align}
Not only it is possible to change $\theta+\varphi$ from angular input $G_f$ but also we can assign $\delta \ge 0$ to converge the desired plane states $\uvec{P}_f=(u_{s,f},v_{s,f})$ in time-domain as
\begin{eqnarray}
\lim\limits_{\uvec{P} \rightarrow \uvec{P}_f}\delta(u_{s},v_{s},u_o,v_o)=0 , \nonumber
\end{eqnarray}
and makes property (\ref{EQ:LiouvilleTheorem}) always true. This is proved under assumption that following condition
\begin{equation}
\sum_{i=1}^{5} \frac{\partial \theta}{\partial x^i}=\sum_{i=1}^{5} \frac{\partial \varphi}{\partial x^i}=0 ,
\label{Eq:Assumptionthetapartialderiv}
\end{equation}
is satisfied. The condition (\ref{Eq:Assumptionthetapartialderiv}) can be derived by using (\ref{Eq:AngleGoalthetaM}) as follows
\begin{eqnarray}
\frac{\partial \theta}{\partial x^i}&=&\frac{\partial\varrho}{\partial x^i}\frac{-1}{1+\varrho^2}  \nonumber\\
&=&\frac{\partial\varrho}{\partial x^i} \frac{-\beta^2_s}{\beta^2_s+ \Big [ \Big(
	\frac{1}{R_{o}}(1-\tan G_f) +\gamma_{s}(-1+\tan
	G_f)-\beta_{s}\tan{G_f} \Big) \Big ]^2}
\end{eqnarray}
where
\begin{equation*}
\varrho=\frac{1}{\beta_{s}}\Big(
\frac{1}{R_{o}}(1-\tan G_f) +\gamma_{s}(-1+\tan
G_f)-\beta_{s}\tan{G_f} \Big).
\end{equation*}
If we choose the control inputs $\{\beta_s,\gamma_s\}$ in a way that the sufficient condition as $1 \ll \varrho^2$ is satisfied, then (\ref{Eq:Assumptionthetapartialderiv}) is true. Since $\varphi$ is a constant value in (\ref{Eq:AngleGoalthetaM}), the relation (\ref{Eq:Assumptionthetapartialderiv}) is satisfied.
Thus, the drift term based on (\ref{EQ:LiouvilleTheorem}) and (\ref{Eq:Assumptionthetapartialderiv}) becomes WPPS. Note that the term $\delta$ is similar to the time-scaling control method that was used for the ball-plate system in \cite{date2004simultaneous}.

\begin{rem}
	The Liouville theorem is a sufficient condition to grant the WPPS property of the drift term. It was proved that a compact orientable manifold i.e., sphere, is Poisson stable since every point on $\mathbb{R}^2$ of topological space is reachable\cite{lobry1974controllability}. Ref. \cite{lian1994controllability} also proved that a Poisson stable dense manifold is equivalent to WPPS.
\end{rem}

In finding the Lie brackets of (\ref{generalsystems}), we have four vector fields in total. In our kinematic model, the spin orientation of the sphere is an important control input that corespondents to $\uvec{g}_3$ vector. Then, the Lie brackets are established as follows
\begin{align*}
\begin{split}
& \uvec{f}=\left[\begin{array}{c}
\sin(\theta+\varphi)\\
\sin(\theta+\varphi)\\
\frac{\sin(\theta+\varphi)[\sin{\psi}-\cos{\psi}]}{R_{o}\cos{v_{o}}}\\
\frac{\sin(\theta+\varphi)[\cos{\psi}+\sin{\psi}]}{R_{o}}\\
\frac{\tan{{v}_{o}}[\sin(\theta+\varphi)(\sin\psi-\cos\psi)+\cos\varphi]}{R_{o}}
\end{array}\right],\uvec{g}_1=\left[\begin{array}{c}
-R_{o}\sin(\theta+\varphi)\\
-R_{o}\sin(\theta+\varphi)\\
\frac{\sin(\theta+\varphi)[\cos{\psi}-\sin{\psi}]}{\cos{v_{o}}}\\
-\sin(\theta+\varphi)[\sin{\psi}+\cos{\psi}]\\
\tan{{v}_{o}}[\sin(\theta+\varphi)(\cos\psi-\sin\psi)]
\end{array}\right],\\
&\uvec{g}_2=\left[\begin{array}{c}
R_{o}\sin(\theta+\varphi)\\
-R_{o}\cos(\theta+\varphi)\\
\frac{-\sin{(\psi+\theta+\varphi)}}{\cos{v_{o}}}\\
-\cos{(\psi+\theta+\varphi)}\\
-\tan{{v}_{o}}\sin{(\psi+\theta+\varphi)}
\end{array}\right],\uvec{g}_3=\left[\begin{array}{c}
0\\
0\\
0\\
0\\
-1
\end{array}\right],\\
\end{split}
\end{align*}
\begin{align}
&[\uvec{f},\uvec{g}_3]= \left[\begin{array}{c}
0\\
0\\
-\frac{\sin(\theta+\varphi)[\cos\psi+\sin\psi]}{R_{o}\cos v_{o}}\\
-\frac{\sin(\theta+\varphi)[\cos\psi-\sin\psi]}{R_{o}}\\
-\frac{\tan v_{o}\sin(\theta+\varphi)[\cos\psi+\sin\psi]}{R_{o}}
\end{array}\right],   [\uvec{f} ,[\uvec{f} ,\uvec{g}_3]]=  \left[\begin{array}{c}
0\\
0\\
\frac{\sin(\theta+\varphi)[-\sin\psi+\cos\psi]}{R_{o}\cos v_{o}}\\
-\frac{\sin(\theta+\varphi)[\cos\psi+\sin\psi]}{R_{o}}\\
\frac{\tan v_{o}\sin(\theta+\varphi)[-\sin\psi+\cos\psi]}{R_{o}}
\end{array}\right].
\end{align}
It is clear that the necessary condition for controllability in Theorem \ref{Controllabilitydrifttheorem} is satisfied since $\dim (\uvec{L_3})= \dim \left\{\uvec{g}_1,\uvec{g}_2,\uvec{g}_3,[\uvec{f},\uvec{g}_3],[\uvec{f},[\uvec{f},\uvec{g}_3]]\right\}=5$, and its determinant is
\begin{align}
&\det \left(\uvec{L_3}\right) =-2  \frac{\sin{v_o}}{\cos^2 v_{o}} \cos\varphi \sin^3(\theta+\varphi) [\sin(\theta+\varphi)+\cos(\theta+\varphi)].
\label{Eq:DeterminantControli}
\end{align}
Note that the determinant (\ref{Eq:DeterminantControli}) has singular points at $\theta+\varphi=k\pi,\;\pi(2k+1)/2$ and $\varphi=\pi(2k+1)/2$ which have to be avoided when we are choosing a desired configuration on the plane. Also, the third singularity is at the local coordinate of the sphere manifold at $v_o=\pi (2k+1)/2$ angles that sends the determinant to infinity and controllability is lost. This can be solved by avoiding the desired angles $v_{o,f}$ near to $\pm\pi/2$ (two points located at two sides of the sphere's equator when $u_o$ is an arbitrary angle in our coordinate system (\ref{Eq:Coorinatequationsphereonplane})).

In this kinematic model all three inputs always \makehighlightn{exist}. In particular, if one of the inputs is removed e.g., $\beta_{s}=0$, the the kinematic model becomes uncontrollable
\begin{align}
\begin{split}
&\dim (\uvec{L})= \dim \big\{\uvec{g}_1,\uvec{g}_3,[\uvec{g}_1,\uvec{g}_3],[\uvec{f},\uvec{g}_3],[\uvec{f},\uvec{g}_1],[\uvec{f},[\uvec{g}_1,\uvec{g}_3]],[\uvec{f},[\uvec{f},\uvec{g}_3]],\\
&[\uvec{f},[\uvec{f},\uvec{g}_1]],[\uvec{g_1},[\uvec{g}_1,\uvec{g}_3]], [\uvec{g_1},[\uvec{f},\uvec{g}_3]],[\uvec{g_1},[\uvec{f},\uvec{g}_1]],[\uvec{g_3},[\uvec{g}_1,\uvec{g}_3]],\\
& [\uvec{g_3},[\uvec{f},\uvec{g}_3]],[\uvec{g_3},[\uvec{f},\uvec{g}_1]], [\uvec{f},[\uvec{f},[\uvec{g}_1,\uvec{g}_3]]],[\uvec{f},[\uvec{f},[\uvec{f},\uvec{g}_3]]],[\uvec{f},[\uvec{f},[\uvec{f},\uvec{g}_1]]],\\
&[\uvec{f},[\uvec{g_1},[\uvec{g}_1,\uvec{g}_3]]], [\uvec{f},[\uvec{g_1},[\uvec{f},\uvec{g}_3]]],[\uvec{f},[\uvec{g_1},[\uvec{f},\uvec{g}_1]]],[\uvec{f},[\uvec{g_3},[\uvec{g}_1,\uvec{g}_3]]], [\uvec{f},[\uvec{g_3},[\uvec{f},\uvec{g}_3]]],\\
&[\uvec{f},[\uvec{g_3},[\uvec{f},\uvec{g}_1]]],[\uvec{f},[\uvec{g_3},[\uvec{f},\uvec{g}_1]]] \big\}=4 \neq 5.
\label{Eq:LieGroupsBeta}
\end{split}
\end{align}
Thus, we define a proposition based on our findings.
\begin{prop}
	All three arc-length-based control inputs, in particular $\gamma_{s}$ and $\beta_{s}$, should always exist in (\ref{EQ:LatestStateEquation}). Otherwise, the rank of Lie bracket for the system with drift is not full (less than five). This makes the kinematic model uncontrollable.
	\label{AllthreeinputsExist}
\end{prop}

\section{Conclusion}
In this paper, we presented a new parametrization based on \makehighlightn{Darboux frame} for path planning and control purposes. First, the kinematic model of the Darboux frame at the contact point between the sphere and the plane was derived. This new formulation is based on a virtual surface sandwiched between surfaces (sphere and plane) to produce arc-length-based control inputs. Next, the induced curvatures of the sphere and plane were developed with relative angular inputs. Then, the Montana kinematic equations with three inputs and five states, that is an underactuated model, was transformed to a fully-actuated one. This \makehighlightn{may facilitate} the control and planning problem for this rolling sphere. In addition, the controllability of the Darboux-frame-based kinematic model was tested. The proposed kinematics preserves the advantages of time- and coordinate-invariance \cite{CuiDarboux2010,DIfgeometry1976} while we have more control inputs to converge the spin-rolling sphere to desired states.
In future research, we plan to establish a path planning strategy with the established Darboux-frame-based kinematics.

\section*{Acknowledgment}
This research was supported, in part, by the Japan Science and Technology Agency, the JST Strategic International Collaborative Research Program, Project No. 18065977.

\section*{References}

\bibliography{mybibfile}

\appendix
\section{ Moving Darboux Frame Preliminaries}
\label{MovingFramePreliminaries}
\begin{figure}[b!]
	\centering	
	\includegraphics[width=2.6 in]{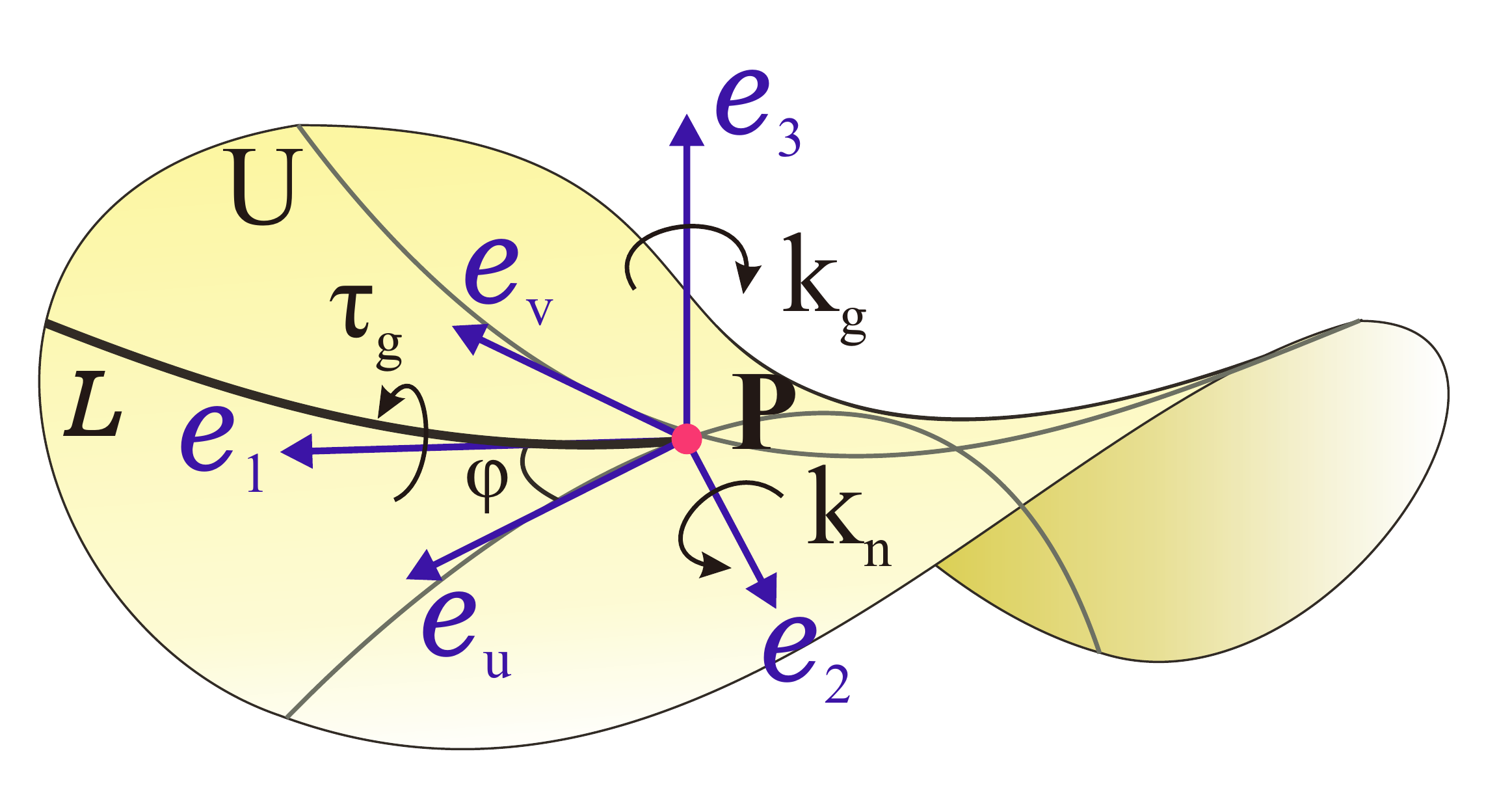}		
	\caption{Two coordinate frames related by $\varphi $ rotational angle about $\bm{e}_3$.}\label{Fig:Simple_Darboux}
\end{figure}
The Darboux frame on an arbitrary surface $U$ is illustrated in Fig.~\ref{Fig:Simple_Darboux}. At the contact point $\uvec{P} \in U$, we introduce unit vectors ($\bm{e}_1,\bm{e}_2,\bm{e}_3$), where $\bm{e}_1$ is a tangent vector to the curve $\uvec{L}$, $\bm{e}_3$ is a normal vector to the $U$ surface and $\bm{e}_2$ is perpendicular to the plane $\bm{e}_3 \times \bm{e}_1$ \cite{Riemannian2002}. The motion of the Darboux frame along the curve $\uvec{L}$ on the surface $U$ is described as \cite{Riemannian2002}
\begin{align}
\begin{split}
&d\uvec{P}=\omega_1^f\bm{e}_1,\\
&d\left[\begin{array}{c}
\bm{e}_1\\
\bm{e}_2\\
\bm{e}_3
\end{array}\right]=\left[\begin{array}{ccc}
0&\omega^f_{12}&\omega^f_{13}\\
-\omega^f_{12}&0&\omega^f_{23}\\
-\omega^f_{13}&-\omega^f_{23}&0
\end{array}\right] \left[\begin{array}{c}
\bm{e}_1\\
\bm{e}_2\\
\bm{e}_3
\end{array}\right].
\end{split}
\label{Eq:GeneralAngularVelDarb}
\end{align}
where $\omega^f_1$, $\omega^f_{12}$, $\omega^f_{23}$ and $\omega^f_{13}$ are the one-forms of the Darboux frame ($\bm{e}_1,\bm{e}_2,\bm{e}_3$). If the curve $\uvec{L}$ is parameterized by the arc length $s$, $\omega^f_1$ is the component of translation of the Darboux frame in the arc-length domain \cite{Riemannian2002}. Also, $\omega^f_{12}$, $\omega^f_{23}$ and $\omega^f_{13}$ are the components of rotation of the Darboux frame in the arc-length domain. The curvature dependencies along the curve $\uvec{L}$ are defined 
by using the following one-form differential relations \cite{CuiDarboux2015}
\begin{align}
k_g=\omega^f_{12}/\omega^f_{1},\;\;k_n=\omega^f_{13}/\omega^f_1,\;\; \tau_g=\omega^f_{23}/\omega^f_1,
\label{EQ:curvaturepropertiesdarboux}
\end{align}
where $k_g$, $k_n$ and $\tau_g$ are, respectively, the geodesic curvature, the normal curvature and the geodesic torsion of the Darboux frame.

Now, introduce the frame ($\bm{e}_u,\bm{e}_v,\bm{e}_3$) induced by the contact coordinates on the surface $U$ and
consider the angle $\varphi$ between the vector $\bm{e}_1$ and the vector $\bm{e}_v$. Then
\begin{eqnarray}
\bm{e}_1&=&\cos\varphi \; \bm{e}_u + \sin{\varphi} \;\bm{e}_v, \nonumber \\
\bm{e}_2&=&-\sin{\varphi}\;\bm{e}_u+ \cos{\varphi} \;\bm{e}_v, \nonumber\\
\bm{e}_3&=& \bm{e}_3,
\end{eqnarray}
Differentiating both sides of this transformation results in
\begin{align}
\left[\begin{array}{c}
\omega^f_{12}\\
\omega^f_{13}\\
\omega^f_{23}
\end{array}\right]=\left[\begin{array}{ccc}
1&0&0\\
0&\cos{\varphi}&\sin{\varphi}\\
0&-\sin{\varphi}&\cos{\varphi}
\end{array}\right]\left[\begin{array}{c}
\omega_{12}\\
\omega_{13}\\
\omega_{23}
\end{array}\right],
\label{Eq:TransformationThetaDar}
\end{align}
where $\omega_{12}$, $\omega_{13}$ and $\omega_{23}$ are the one-forms for the induced coordinates on $U$.
We will use these relationships to develop curvature equations between the Darboux frame and the frame induced by the contact coordinates on $U$ in \ref{DarbouxFrameKinematicsSec}.

\section{Darboux-Frame-Based Kinematics Between Two Surfaces with Sandwiched Virtual Surface}
\label{VirtualSurfaceDarbouxSurfaces}
\begin{figure}[b!]
	\centering	
	\includegraphics[width=2.7 in]{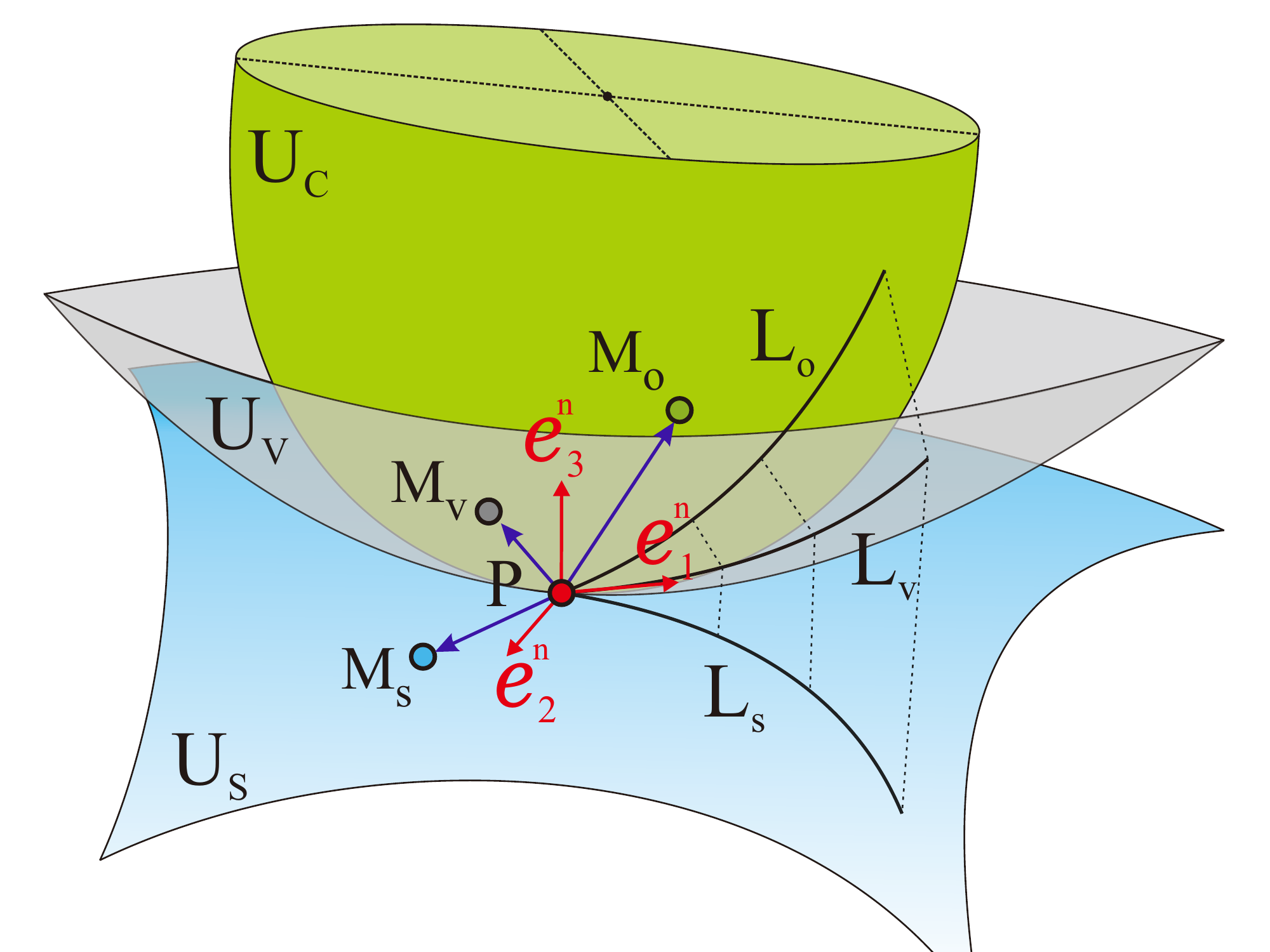}		
	\caption{Rotating object $U_C$ on the fixed surface $U_S$ with sandwiched virtual surface $U_V$. Note that $n$ in $\{\bm{e}^n_1,\bm{e}^n_2,\bm{e}^n_3\}$ frame stands for fixed surface (plane) $s$,  virtual surface $v$ and rotating object (sphere) $o$ at each of these compact surfaces.}\label{Fig:VirtualSurface}
\end{figure}
The induced curvature (differential characteristics) of a virtual surface (\ref{EQ:TheCurvaturerelevantDif}) sandwiched between the rotating object (sphere) and the fixed surface (plane) are derived here.
We conduct the derivation by using the results of the study on the induced curvatures between two surfaces \cite{CuiDarboux2010}. Here, we introduce a  virtual surface that is sandwiched between two surfaces (sphere and plane in our case). This virtual surface relates the arc-length-based inputs with derived curvatures in (\ref{Eq:MainCurvatureDarobux}) to manipulate the spin-rolling sphere angular coordinates.

Consider an object $U_C$ spin-rolling on the surface $U_S$.  A virtual moving surface $U_V$ is sandwiched as shown in Fig.~\ref{Fig:VirtualSurface}. As explained earlier, the Darboux frames ($\bm{e}^n_1,\bm{e}^n_2,\bm{e}^n_3$) of all three surfaces are aligned with each other (see Fig. \ref{Fig:VirtualSurface}). The spin-rolling object (sphere), the fixed surface (plane) and the virtual surface keep the same contact point $\uvec{P}$. The motion of the Darboux frame at the contact point $\uvec{P}$ contains is described as
\begin{align}
\begin{split}
& \frac{d\uvec{P}}{ds}=\bm{e}_1,\\
&d\left[\begin{array}{c}
\bm{e}_1\\
\bm{e}_2\\
\bm{e}_3
\end{array}\right]=\left[\begin{array}{ccc}
0&k_g&k_n\\
-k_g&0&\tau_g\\
-k_n&-\tau_g&0
\end{array}\right]\left[\begin{array}{c}
\bm{e}_1\\
\bm{e}_2\\
\bm{e}_3
\end{array}\right],\;\;\;\;\;\;\;\;\;\;\;\;
\end{split}
\label{Eq:DArbouxFrame}
\end{align}
where, regardless of coordinate dependency, $s$ is the arc length of the curve $\uvec{L}_{s}$, and $k_g$, $k_n$ and $\tau_g$ are the geodesic curvature, the normal curvature and the geodesic torsion of the Darboux frame. The right-handed orthonormal unit vectors are aligned in same direction for all objects with respective subscripts, for example, the object $U_C$ has $(\bm{e}^{o}_1,\bm{e}^{o}_2 ,\bm{e}^{o}_3)$ (see Fig. \ref{Fig:VirtualSurface}). Firstly, the position of an arbitrary fixed point $\uvec{M}_{o}$ on body $U_C$ is set as
\begin{equation}
\uvec{M}_{o}=\uvec{P}+\sigma^{o}_1\bm{e}^{o}_1 +\sigma^{o}_2\bm{e}^{o}_2 +\sigma^{o}_3\bm{e}^{o}_3,
\label{Eq:positionmatrixdarbouxsphere}
\end{equation}
where $\sigma$ stands for a scalar coordinate of the considered object, here $\{\sigma^{o}_1,\sigma^{o}_2,\sigma^{o}_3\}$ is for the rolling sphere. Differentiating (\ref{Eq:positionmatrixdarbouxsphere}) with respect to the arc length $s$ and using (\ref{Eq:DArbouxFrame}) gives
\begin{equation}
\begin{split}
&\frac{d\uvec{M}_{o}}{ds}= \left[\begin{array}{c}
1+ d\sigma^{o}_1/ds-\sigma^{o}_2k^{o}_g-\sigma^{o}_3k^{o}_n \\
d\sigma^{o}_2/ds +\sigma^{o}_1k^{o}_g-\sigma^{o}_3\tau^{o}_n\\
d\sigma^{o}_3/ds +\sigma^{o}_1k^{o}_n+\sigma^{o}_2\tau^{o}_n
\end{array}\right]^T\left[\begin{array}{c}
\bm{e}^{o}_1 \\
\bm{e}^{o}_2 \\
\bm{e}^{o}_3
\end{array}\right].\\
\end{split}
\label{Eq:TheDifferentiatedPosi}
\end{equation}
Since vector $\uvec{M}_o$ is fixed, the derivative with respect to the arc length is $ d\uvec{M}_{o}/ds=0$. This is true for all three surfaces. Therefore, for the rotating surface $U_C$, we have
\begin{align}
\begin{split}
&d\sigma^{o}_1/ds= \sigma^{o}_2k^{o}_g+\sigma^{o}_3k^{o}_n-1,\;d\sigma^{o}_2/ds=-\sigma^{o}_1k^{o}_g\\
&+\sigma^{o}_3\tau^{o}_n,\;d\sigma^{o}_3/ds=-\sigma^{o}_1k^{o}_n-\sigma^{o}_2\tau^{o}_n.
\label{Eq:RESULTOFEQUATIn2}
\end{split}
\end{align}
The same reasoning can be applied for the point $\uvec{M}_v$ on surface $U_V$ with the arc length of $s'$,
\begin{equation}
\begin{split}
&\frac{d\uvec{M}_{v}}{ds}= \left[\begin{array}{c}
1+ d\sigma^{v}_1/ds'-\sigma^{v}_2k^{v}_g-\sigma^{v}_3k^{v}_n \\
d\sigma^{v}_2/ds' +\sigma^{v}_1k^{v}_g-\sigma^{v}_3\tau^{v}_n\\
d\sigma^{v}_3/ds' +\sigma^{v}_1k^{v}_n+\sigma^{v}_2\tau^{v}_n
\end{array}\right]^T\left[\begin{array}{c}
\bm{e}^{v}_1\\
\bm{e}^{v}_2\\
\bm{e}^{v}_3
\end{array}\right],\\
\end{split}
\label{Eq:TheDifferentiatedPosiVitrual}
\end{equation}
which results in
\begin{align}
\begin{split}
&d\sigma^{v}_1/ds'= \sigma^{v}_2k^{v}_g+\sigma^{v}_3k^{v}_n-1,\;d\sigma^{v}_2/ds'=\\
&-\sigma^{v}_1k^{v}_g+\sigma^{v}_3\tau^{v}_n,\;d\sigma^{v}_3/ds'=-\sigma^{v}_1k^{v}_n-\sigma^{v}_2\tau^{v}_n.
\end{split}
\end{align}
Note that the curves $\uvec{L}_{o}$ and $\uvec{L}_{v}$  are traversed with the same velocity.
Since arc lengths of the two curves over the same time period are the same, we have  $\sigma^{o}_q=\sigma^{v}_q$ and $\sigma_q^{o}/ds=\sigma_q^{v}/ds'$ for $q\in[1,3]$. The substitution of (\ref{Eq:RESULTOFEQUATIn2}) into (\ref{Eq:TheDifferentiatedPosiVitrual}), gives
\begin{align}
\begin{split}
&\frac{d\uvec{M}_{v}}{ds}= \left[\begin{array}{c}
\sigma^{v}_2k^{ov}_g+\sigma^{v}_3k^{ov}_n\\
-\sigma^{v}_1k^{ov}_g+\sigma^{v}_3\tau^{ov}_g\\
-\sigma^{v}_1k^{ov}_n-\sigma^{v}_3\tau^{ov}_g
\end{array}\right]^T\left[\begin{array}{c}
\bm{e}^{v}_1\\
\bm{e}^{v}_2\\
\bm{e}^{v}_3
\end{array}\right],\\
\end{split}
\label{Eq:TheDifferentiatedPosiVitruallast}
\end{align}
where $k^{ov}_g=k^{o}_g-k^{v}_g$, $k^{ov}_n=k^{o}_n-k^{v}_n$ and $\tau^{ov}_g=\tau_g^{o}-\tau^{v}_g$. Next, the same reasoning is applied to the virtual surface $U_V$ with respect to fixed surface $U_S$ with curves $\uvec{L}_v$ and $\uvec{L}_s$, and this results in
\begin{align}
\begin{split}
&\frac{d\uvec{M}_{s}}{ds}= \left[\begin{array}{c}
\sigma^{s}_2k^{vs}_g+\sigma^{s}_3k^{vs}_n\\
-\sigma^{s}_1k^{vs}_g+\sigma^{s}_3\tau^{vs}_g\\
-\sigma^{s}_1k^{vs}_n-\sigma^{s}_3\tau^{vs}_g
\end{array}\right]^T\left[\begin{array}{c}
\bm{e}^{s}_1\\
\bm{e}^{s}_2\\
\bm{e}^{s}_3
\end{array}\right],\\
\end{split}
\label{Eq:VirtualandSurfaceGround}
\end{align}
where $k^{vs}_g=k^{v}_g-k^{s}_g$, $k^{vs}_n=k^{v}_n-k^{s}_n$ and $\tau^{vs}_g=\tau_g^{v}-\tau^{s}_g$. In (\ref{Eq:VirtualandSurfaceGround}), we rearrange the curvature properties as
\begin{align}
\begin{split}
k^{v}_g= k^{vs}_g+k^{s}_g,\;k^{v}_n=k^{vs}_n+k^{s}_n,\tau_g^{v}=\tau^{vs}_g+\tau^{s}_g.
\end{split}
\label{Eq:Reordireing}
\end{align}
Substituting (\ref{Eq:Reordireing}) back to (\ref{Eq:TheDifferentiatedPosiVitruallast}) and assuming $k^{vs}_g=\alpha_s$, $k^{vs}_n=\gamma_{s}$,  $\tau^{vs}_g=\beta_{s}$, gives the angular velocity of the Darboux frame at point $\uvec{P}$
\begin{equation}
\bm{\omega}^*=\delta(-\tau^{*}_g \bm{e}_1+k^{*}_n\bm{e}_2-k^{*}_g\bm{e}_3),
\label{Eq:LASTFORMAT}
\end{equation}
where
\begin{align}
\begin{split}
&k^{*}_g= k^{o}_g-k^{s}_g-\alpha_s , k^{*}_n=k^{o}_n-k^{s}_n-\gamma_{s},\tau^{*}_g=\tau_g^{o}-\tau^{s}_g-\beta_{s}.
\label{Eq:Curvatureproperywithvirtual}
\end{split}
\end{align}
The relationships (\ref{Eq:Curvatureproperywithvirtual}) show the parameterized differential characteristics of the virtual surface in relation to the induced curvatures of the moving object and fixed surface. In our case, the differential characteristics of the sphere are subtracted from the virtual surface to create the corresponding angular velocity (\ref{Eq:LASTFORMAT}). \makehighlightn{The Darboux frame angular velocity (\ref{Eq:LASTFORMAT}) for two surfaces without virtual surface were presented in \cite{CuiDarboux2015,cui2020sliding}}. Note that changes in the virtual surface's curvatures are projected on both the sphere and the plane traveling paths, $\uvec{L}_{o}$ and $\uvec{L}_{s}$. This gives the physical meaning of the arc-length-based inputs $\{\gamma_{s}$, $\alpha_s$, $\beta_{s}\}$ in (\ref{Eq:MainCurvatureDarobuxCS}).

\section{The Induced Curvatures Variation in a Given Direction}
\label{DarbouxFrameKinematicsSec}
We derive the relation between angular rotation of the Darboux frame about $\bm{e}_3$ (see Fig.~\ref{Fig:Simple_Darboux}) and the normal curvature, geodesic torsion and geodesic curvature on an arbitrary surface. Let $f_l(u,v)$ be a differentiable manifold in $S\subset \mathbb{R}^3$. The unit vectors $\bm{e}_v$ and $\bm{e}_u$ are along $v$ and $u$ curves on a moving point $\uvec{P} \subset S$. The differentiated map of the point $\uvec{P}$ is \cite{Riemannian2002}
\begin{align*}
d\uvec{P}=\uvec{r}_u du+ \uvec{r}_v dv= \omega_{1} \bm{e}_u + \omega_{2} \bm{e}_v,
\end{align*}
hence, by defining $\omega_{1}=\sqrt{E}du,\omega_{2}=\sqrt{G}du$, we have
\begin{align}
\left[\begin{array}{c}
\uvec{r}_u\\
\uvec{r}_v\\
\varLambda
\end{array}\right]=\left[\begin{array}{ccc}
\sqrt{E}&0&0\\
0&\sqrt{G}&0\\
0&0&1
\end{array}\right]\left[\begin{array}{c}
\bm{e}_u\\
\bm{e}_v\\
\bm{e}_3
\end{array}\right]=A \left[\begin{array}{c}
\bm{e}_u\\
\bm{e}_v\\
\bm{e}_3
\end{array}\right],
\label{Eq:GnEquationfirstfundemental}
\end{align}
where $E=\uvec{r}_u\cdot \uvec{r}_u$, $G=\uvec{r}_v \cdot \uvec{r}_v$ and $\varLambda$ are the coefficients for the first fundamental form and normal vector of the surface $S$, respectively.
Differentiating the left side of Eq. (\ref{Eq:GnEquationfirstfundemental}) results in
\begin{equation}\small{
	\begin{split}
	&d\left[\begin{array}{c}
	\uvec{r}_u\\
	\uvec{r}_v\\
	\varLambda
	\end{array}\right]=du\left[\begin{array}{c}
	\uvec{r}_{uu}\\
	\uvec{r}_{uv}\\
	\varLambda_{u}
	\end{array}\right]+dv\left[\begin{array}{c}
	\uvec{r}_{uv}\\
	\uvec{r}_{vv}\\
	\varLambda_{v}
	\end{array}\right]=\\
	& \bigg(du\left[\begin{array}{ccc}
	\varGamma^1_{11}&\varGamma^2_{11}&L\\
	\varGamma^1_{12}&\varGamma^2_{12}&M\\
	W^1_1&W^2_1&0
	\end{array}\right]+dv\left[\begin{array}{ccc}
	\varGamma^1_{12}&\varGamma^2_{12}&M\\
	\varGamma^1_{22}&\varGamma^2_{22}&N\\
	W^1_2&W^2_1&0
	\end{array}\right]\bigg)
	\left[\begin{array}{c}
	\uvec{r}_u\\
	\uvec{r}_v\\
	\varLambda
	\end{array}\right]
	\end{split}}
\label{Eq:lastangular11}
\end{equation}
where $\varGamma^k_{ij}$, $W^j_i$, $L$, $M$ and $N$ are coefficients of Gauss and Weingarten equations, and the rest of the three coefficients are the second fundamental form of surface, which are \cite{DIfgeometry1976}
\begin{align*}
\begin{split}
&\varGamma^1_{11}=\frac{GE_u-2FF_u+FE_v}{2(EG-F^2)},\; \varGamma^2_{11}=\frac{2EF_u-EE_v+FE_u}{2(EG-F^2)} \\
&\varGamma^1_{12}=\frac{GE_v-FG_u}{2(EG-F^2)},\; \varGamma^2_{12}=\frac{EG_u-FE_v}{2(EG-F^2)}, \\
& \varGamma^1_{22}=\frac{2GF_v-2GG_u-FG_v}{2(EG-F^2)},\;\varGamma^2_{22}=\frac{EG_v-2FF_v+FG_u}{2(EG-F^2)},\\
&W^1_1=\frac{MF-LG}{EG-F^2},W^2_1=\frac{LF-ME}{EG-F^2},\; W^1_2=\frac{NF-MG}{EG-F^2}, W^2_2=\frac{MF-NE}{EG-F^2}.\\
\end{split}
\end{align*}
Next, differentiating the right-hand side of (\ref{Eq:GnEquationfirstfundemental}) gives
\begin{align}
\begin{split}
&d \Bigg( \uvec{B} \left[\begin{array}{c}
\bm{e}_u\\
\bm{e}_v\\
\bm{e}_3
\end{array}\right]\Bigg ) = \Bigg(d\uvec{B}\left[\begin{array}{c}
\bm{e}_u\\
\bm{e}_v\\
\bm{e}_3
\end{array}\right]+\uvec{B} d \Bigg(\left[\begin{array}{c}
\bm{e}_u\\
\bm{e}_v\\
\bm{e}_3
\end{array}\right] \Bigg ) \Bigg )
\\
&=\Bigg(d \uvec{B}+\uvec{B}
\left[\begin{array}{ccc}
0&\omega_{12}&\omega_{13}\\
-\omega_{12}&0&\omega_{23}\\
-\omega_{13}&-\omega_{23}&0
\end{array}\right]
\Bigg )\left[\begin{array}{c}
\bm{e}_u\\
\bm{e}_v\\
\bm{e}_3
\end{array}\right].
\end{split}
\label{Eq:lastangular12}
\end{align}
where $\omega_{12}$, $\omega_{13}$ and $\omega_{23}$ are the one-forms for ($\bm{e}_u,\bm{e}_v,\bm{e}_3$) here. Also,  (\ref{Eq:lastangular11}-\ref{Eq:lastangular12}) yield the angular velocities of rotating body $(\bm{e}_u,\bm{e}_v,\bm{e}_3)$ for isometric surfaces
(under assumption of $F=0$)
\begin{align}
\begin{split}
&\omega_{12}=\frac{-E_vdu+G_udv}{2\sqrt{EG}},\; \omega_{13}=\frac{Ldu+Mdv}{\sqrt{E}},\;\omega_{23}=\frac{Mdu+Ndv}{\sqrt{G}}.
\end{split}
\end{align}
Next, the differential characteristics (\ref{EQ:curvaturepropertiesdarboux}) are calculated for $u$ and $v$ coordinates.
For each principle, the derivative of  perpendicular curve becomes zero,
For the $u$-curve we have
\begin{align}
\begin{split}
&k_{gu}=\frac{\omega_{12}}{\omega_{1}}=\frac{-E_vdu}{2\sqrt{EG}} \cdot \frac{1}{\sqrt{E}du}=-\frac{E_v}{2E\sqrt{G}},\\
&k_{nu}=\frac{\omega_{13}}{\omega_{1}}=\frac{Ldu}{\sqrt{E}} \cdot  \frac{1}{\sqrt{E}du}=\frac{L}{E},\\
&\tau_{gu}=\frac{\omega_{23}}{\omega_1}=\frac{Mdu}{\sqrt{G}} \cdot \frac{1}{\sqrt{E}du}=\frac{M}{\sqrt{EG}},
\label{Eq:Curvatureinduceducurve}
\end{split}
\end{align}
and for the $v$-curve we have
\begin{align}
\begin{split}
k_{gv}=\frac{G_u}{2G\sqrt{E}},\;
k_{nv}=\frac{N}{G},\;
\tau_{gv}=-\frac{M}{\sqrt{EG}}.
\end{split}
\label{Eq:Curvatureinducedvcurve}
\end{align}
It should be note here that the frame on the $v$-curve is set as ($\bm{e}_v,-\bm{e}_u,\bm{e}_3$) and the angle $\varphi$ in (\ref{Eq:TransformationThetaDar}) equals to $\pi/2$. Finally, the differential characteristics in a given direction of the curve $\uvec{L}_{s}$,
derived in dependence on the arc length $s$ \cite{DIfgeometry1976}, are obtained by using  (\ref{Eq:GeneralAngularVelDarb}), (\ref{Eq:TransformationThetaDar}) and (\ref{Eq:Curvatureinduceducurve}-\ref{Eq:Curvatureinducedvcurve})
\begin{align}
\begin{split}
&k_n=\frac{\omega^{f}_{13}}{ds}=\frac{\omega_{13}\cos\varphi+\omega_{23}\sin\varphi}{ds}=k_{nu}\cos^2\varphi+2\tau_{gu}\cos\varphi\sin\varphi+k_{nv}\sin^2\varphi,\\
&\tau_g=\frac{\omega^{f}_{23}}{ds}=\frac{-\omega_{13}\sin\varphi+\omega_{23}\cos\varphi}{ds}=\tau_{gu}\cos{2\varphi}+\frac{1}{2}(k_{nv}-k_{nu})\sin 2\varphi,\\
&k_g=\frac{\omega^{f}_{12}}{ds}=\frac{\omega_{12}}{ds}=k_{gu}\cos\varphi+k_{gv}\sin\varphi,
\end{split}
\label{Eq:MainCurvatureDarobux}
\end{align}
Note that formulas (\ref{Eq:MainCurvatureDarobux}) are designed for the induced curvatures are changing depending on the angle $\varphi$
(the angle of rotation of the Darboux frame along with $\bm{e}_1$ as shown in Fig.~\ref{Fig:Simple_Darboux}). \makehighlightn{A similar derivation for (\ref{Eq:MainCurvatureDarobux}) with a spin rate variable was shown in earlier work of L. Cui and J. Dai  \cite{CuiDarboux2015}.}

\section{Parametrization Limitation of the Montana Kinematics}
\label{Sec:Montanalimitationproblem}
The Montana kinematic equations \cite{Montana1988} give a simple relationship between the derivatives of the contact coordinates and the angular velocity. However, they have limitations in assigning extra angular inputs.
Let us assume that we want to constrain spin-rolling motion of the sphere on the plane trajectory $\uvec{L}_s$ by an arbitrary angle.
This angle can be assigned to the considered curve on the plane $\uvec{L}_s$ while the sphere curve $\uvec{L}_o$ is controlled through the angular velocity based on a desired motion planning strategy. This is similar to the 2-DoF mobile robots e.g., cars, where the kinematic models have angular velocity and steering angle inputs. The inputs components of the plane states $\{{u}_s,{v}_s\}$ can be found from differential equations  (\ref{Eq:LastMonata2D}) with the following definition
\begin{equation}
\begin{split}
&G_f \overset{\Delta}{=} \tan^{-1}\left(\frac{\dot{v}_{s}}{\dot{u}_{s}}\right)=\tan^{-1}\left(\frac{d v_{s}}{d u_{s}}\right)=\tan^{-1}\Bigg ( \frac{-R_o \omega^o_x}{R_o \omega^o_y}\Bigg ),
\label{Eq:GfMontanaKienatmics}
\end{split}
\end{equation}
where the angle $G_f$ can be set arbitrarily. With the use of (\ref{Eq:GfMontanaKienatmics}), we have
\begin{equation}
\omega^o_x=-\omega^o_y\tan G_f .
\label{Eq:omegaGf}
\end{equation}
By substituting (\ref{Eq:omegaGf}) into Montana kinematic equations (\ref{Eq:LastMonata2D}), we have
\begin{equation}
\begin{split}
\left[\begin{array}{c}
\dot{u}_{s}(t)\\
\dot{v}_{s}(t)\\
\dot{u}_{o}(t)\\
\dot{v}_{o}(t)\\
\dot{\psi}(t)
\end{array}\right]=\left[\begin{array}{ccc}
0& R_{o}&0\\
- R_{o}&0&0\\
-\sin{\psi}/\cos{{v}_{o}}&-\cos{\psi}/\cos{{v}_{o}}&0\\
-\cos{\psi}&\sin{\psi}&0\\
-\sin{\psi}\tan{{v}_{o}}&-\cos{\psi}\tan{{v}_{o}} &-1
\end{array}\right]\left[\begin{array}{c}
-\omega^{o}_y\tan G_f \\
\omega^{o}_y\\
\omega^{o}_z
\end{array}\right].\\
\end{split}
\label{Eq:LastMonata2DConstraint}
\end{equation}
Although $G_f$ changes the angular direction on the plane for achieving the desired plane states on the curve $\uvec{L}_s$ with imposed constraint ($G_f$), we lose the input $\omega^o_x$ and the rest of the remaining states in the system $\{u_o,v_o,\psi\}$ should be controlled with only 2 inputs $\{\omega^o_y,\omega^o_z\}$. This makes it harder to develop control strategies by using Montana kinematic equations \makehighlightn{(five states and three inputs)} only. We can see that the proposed Darboux-frame-based kinematics (\ref{EQ:LatestStateEquation}) does not have this limitation because it has three arc-length-based inputs including the constraining angular input $G_f$.

\end{document}